\pdfoutput=1

\documentclass[11pt]{article}

\usepackage[final]{acl}

\usepackage{times}
\usepackage{enumitem}
\usepackage{latexsym}

\usepackage[T1]{fontenc}

\usepackage[utf8]{inputenc}

\usepackage{microtype}

\usepackage{inconsolata}
\usepackage[normalem]{ulem}
\useunder{\uline}{\ul}{}
\usepackage{graphicx}
\usepackage[normalem]{ulem}
\usepackage{booktabs}
\usepackage{multirow}
\usepackage{algorithm}
\usepackage{algpseudocode}
\usepackage{amsmath}
\usepackage{amssymb}
\usepackage{subcaption}
\usepackage{textcomp}

\newcommand{\eg}{e.g.,}
\newcommand{\ie}{i.e.,}
\newcommand{\etc}{etc.}

\newcommand\figref[1]{Figure~\ref{#1}}

\newcommand\tabref[1]{Table~\ref{#1}}

\newcommand\secref[1]{\S\ref{#1}}

\newcommand\appref[1]{Appendix~\ref{#1}}

\newcommand{\dataset}[1]{\texttt{HACo-Det}}

\newcommand{\fakeparagraph}[1]{\vspace{1mm}\noindent\textbf{#1}}

\title{\dataset{}: A Study Towards Fine-Grained Machine-Generated Text Detection under Human-AI Coauthoring}

\author{
 \textbf{Zhixiong Su\thanks{~Equal contribution}\textsuperscript{1}}\hspace{0.33cm}
 \textbf{Yichen Wang\footnotemark[1]\textsuperscript{1, 2}}\hspace{0.33cm}
 \textbf{Herun Wan\textsuperscript{1}}\hspace{0.33cm}
 \textbf{Zhaohan Zhang\textsuperscript{3}}\hspace{0.33cm}
 \textbf{Minnan Luo\thanks{~Corresponding author: Minnan Luo, School of Computer Science and Technology,
Xi’an Jiaotong University, Xi’an 710049, China}\textsuperscript{1}}
\\
 \textsuperscript{1}Xi'an Jiaotong University \hspace{0.33cm}
 \textsuperscript{2}University of Chicago \hspace{0.33cm}
 \textsuperscript{3}Queen Mary University of London
\\
\texttt{\{15958057,wanherun\}@stu.xjtu.edu.cn} \hspace{0.5cm}
\texttt{yichenzw@uchicago.edu} \\  \texttt{zhaohan.zhang@qmul.ac.uk} \hspace{0.5cm}
\texttt{minnluo@xjtu.edu.cn}
}

\begin{document}
\maketitle
\begin{abstract}
The misuse of large language models (LLMs) poses potential risks, motivating the development of machine-generated text (MGT) detection.
Existing literature primarily concentrates on binary, document-level detection, thereby neglecting texts that are composed jointly by human and LLM contributions.
Hence, this paper explores the possibility of fine-grained MGT detection under human-AI coauthoring.
We suggest fine-grained detectors can pave pathways toward coauthored text detection with a numeric AI ratio.
Specifically, we propose a dataset, \dataset{}, which produces human-AI coauthored texts via an automatic pipeline with word-level attribution labels.
We retrofit seven prevailing document-level detectors to generalize them to word-level detection.
Then we evaluate these detectors on \dataset{} on both word- and sentence-level detection tasks.
Empirical results show that metric-based methods struggle to conduct fine-grained detection with a 0.462 average F1 score, while finetuned models show superior performance and better generalization across domains. 
However, we argue that fine-grained co-authored text detection is far from solved.
We further analyze factors influencing performance, \eg{} context window, and highlight the limitations of current methods, pointing to potential avenues for improvement.
\end{abstract}

\section{Introduction}
The generative capability of large language models (LLMs), \eg{} GPT-4o \citep{openai2024gpt4o}, Llama-3 \citep{llama3modelcard}, and Claude-3.5-Sonnet \citep{anthropic2024claude35s}, has been rapidly developed and gain further capabilities of instruction-following \citep{qin2024infobench} and interaction \citep{castillo2024beyond}.
It leads to the emergence and prevalence of human-AI interactive and collaborative generation systems, \eg{} GPT-4o-canvas \citep{openai2024canvas}, Notion \citep{notion2023ai}, and Wordcraft \citep{google2024wordcraft}. 
However, this raises broader public concerns about its misuse, \eg{} academic plagiarism \citep{Jarrah2023UsingCI}, privacy leak \citep{mireshghallah2022quantifying}, and hallucination \citep{Li2023HaluEvalAL} \etc{}, hence spurring the creation of machine-generated text (MGT) detection \citep{gehrmann-etal-2019-gltr, zellers2019defending, mitchell2023detectgpt}.
Literature of MGT detection usually assigns a binary label, \ie{} either human-written text (HWT) or MGT, on the document level \cite{mitchell2023detectgpt, liu2023coco, hu2023radar, wang2024stumbling, dugan-etal-2024-raid, li2025iron}.
Among these MGT detection methods are two predominant categories:
metric-based methods compute numeric metrics, e.g., logits, of texts in a white-box setting;
finetune-based methods train classification models on annotated corpora (details at \secref{sec:exp_set} and \appref{sec:appendix_baseline}).

However, the binary document-level MGT detection task and methods might be inadequate for the current rising trend of human-AI collaboration \citep{lee2022coauthor, chakrabarty2022help, reza2024abscribe, li2024value, shu2024rewritelm, reza2025co}. 
It further calls for a step toward fine-grained MGT detection to attribute partial authorship to humans or machines.
Heated debates have occurred on the dilemma of authorship attribution of human-AI coauthored contexts for binary document-level setting \citep{tripto2023ship}.
Fine-grained detection is a potential way to mitigate the controversy, meanwhile reaching an interpretable prediction with a numeric AI ratio and localization \citep{gehrmann-etal-2019-gltr,kushnareva2023artificial}. 
Different forms of fine-grained tasks, e.g., triple classification with Mixtext class \citep{gao2024llm}, boundary detection \citep{kushnareva2023artificial}, and MGT localization \citep{zhang2024machine}, are proposed. 


However, some blind spots still exist in MGT detection under coauthoring.
Firstly, some datasets \citep{gehrmann-etal-2019-gltr, hu2023radar, liu2023coco, pu2023zero} utilize a human-written beginning as the prompt and request LLMs to continue. They label these outputs as MGT, but we question that these are partly human-written.
Secondly, some works use paraphrasing as the machine generation process and directly label fine-grained attribution of paraphrased contexts to the machine \citep{zhang-etal-2024-llm, li2024spotting}.
We argue that, if portions of contents remain lexically unchanged during paraphrasing, their authorship should be attributed to the original author. 
Moreover, \citet{kushnareva2023artificial, gao2024llm} stop at single-turn collaboration, but realistic interactions could be more complex.
A better task design is to be explored. 

In this paper, we propose a novel task and benchmark for human-AI coauthored text detection, identifying their fine-grained authorship at the word level, and aggregating to the sentence level (\secref{sec:data_task}). 
We propose a novel dataset, \dataset{}, utilizing dominant instruct LLMs to partially paraphrase texts in multiple turns (\secref{sec:data}). 
Furthermore, we comprehensively reform the current detectors (\secref{sec:exp_set}) to our tasks. 
In \secref{sec:result}, we evaluate their in-domain and out-of-domain performance at both word and sentence levels. 
Further in \secref{sec:ana}, we analyze the limitations of current detectors, including capability in generalization and zero-shot prediction. We propose possible trails toward improvement, \eg{} larger context windows, augment training corpus diversity, \etc{} 
In summary, our work provides three main contributions:
\begin{itemize}
    \item We propose a word-level labeled MGT detection dataset, \dataset{}, in which texts are collaboratively generated by humans and predominantly instruct LLM. 
    \item We reform seven MGT detectors to our tasks and evaluate them on \dataset{} at word and sentence levels, finding large space exists to improve current methods, especially metric-based ones.
    \item We further analyze the limitation of current detectors and influencing factors, \eg{} context windows, suggesting some possible ways for enhancement. 
\end{itemize}

\begin{figure*}[t]
  \centering
  \includegraphics[width=0.7\textwidth]{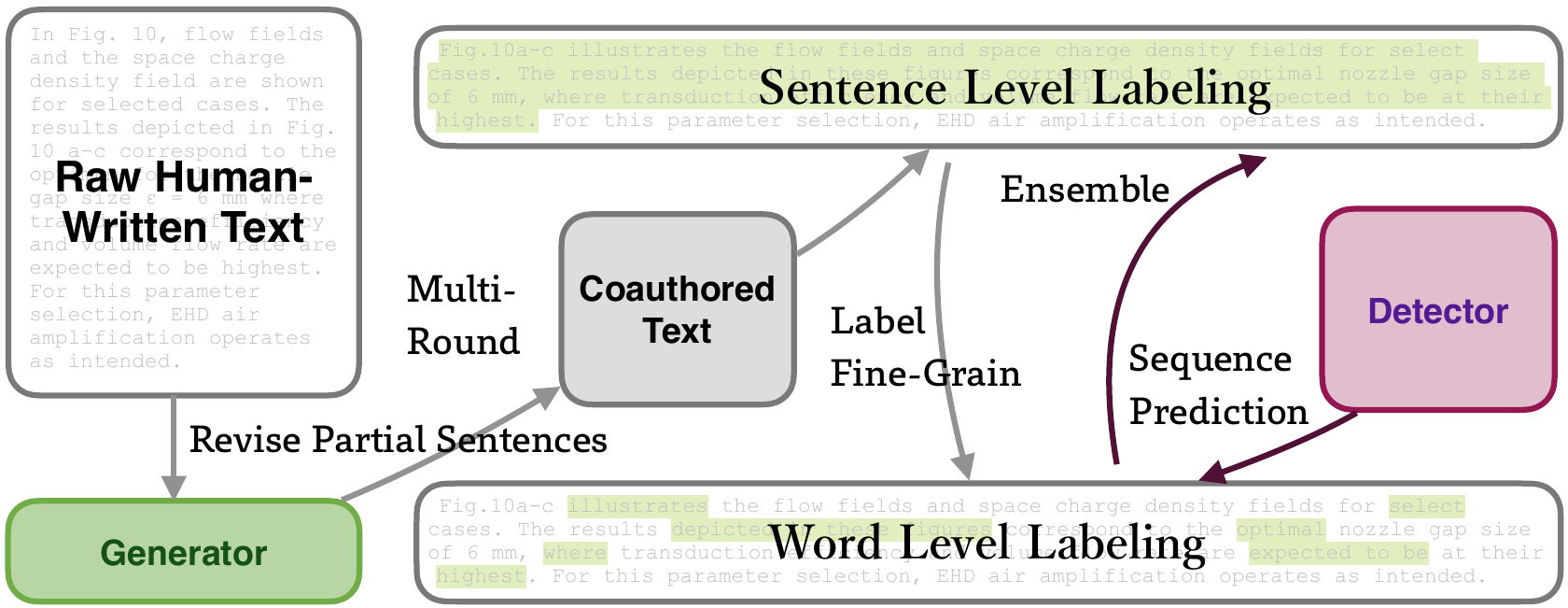}
  \caption{An overview of the study pipeline in \dataset{}. Firstly, we sample texts according to a curated rule from the raw human-written texts. Then we use instruct LLM to paraphrase them multiple rounds. We label the texts at the word level and sentence level, according to the detection task setting (\secref{sec:data_const}). In the main experiment (\secref{sec:exp_set}), the detectors do sequence prediction at the word level, then ensemble the results to the sentence level. }
  \label{fig:pip_study}
\end{figure*}


\section{Related Works}

\subsection{Document-Level MGT Detection}
Previous works on MGT detection treat the problem as a binary classification task with document-wise labels.
The goal is to find discriminative metrics or representations of the input document.
For example, LogLikelihood \cite{solaiman2019release}, Entropy \citep{gehrmann-etal-2019-gltr}, Rank \cite{gehrmann-etal-2019-gltr}, and Log-Rank \cite{mitchell2023detectgpt} are found to be promising metrics in differentiating MGT and HWT.
Given the convenience of obtaining metrics from the output distribution of the model, it holds a strong assumption that the generators are white-box, which is not practical in closed-source models \citep{achiam2023gpt, anthropic2024claude35s}.
Some works train a finetune-based detector to extract detectable representations from texts.
\citet{guo2023close} use RoBERTa \cite{liu2019roberta} as detector backbone for acquiring text embedding and classification.
\citet{liu2023coco} incorporates the coherence graph into the text representation and further \citet{liu2024does} use perturbed text as additional input for learning robust and discriminative embeddings for classification.
\citet{hu2023radar, li2025iron} apply adversarial training to improve the robustness of the MGT detectors against attacks.
Different from the problem setting in the previous works, we model MGT detection as a sequence-to-sequence classification task where we label every word as MGT or HWT to identify the MGT part in hybrid documents.

\subsection{Fine-Grained MGT Detection}

Recent works try to formalize more fine-grained detection tasks than document-level prediction.
Fine-grained MGT detection shows the potential to handle coauthored text detection and provide interpretable prediction, e.g., numeric AI ratio and localization.
\citet{gao2024llm} propose a triple classification setting, introducing a new class {Mixtext}, in which humans and LLM are both involved in the generation process.
\citet{kushnareva2024ai} introduce a boundary detection task, under which the authorship of each text is started with humans and shifted to machines at a specific boundary position.
\citet{li-etal-2024-spotting} propose a detection framework aiming to identify paraphrased text spans by sentence-level classification.
\citet{zhang2024machine} propose the MGT localization task to detect short MGT sentence spans and a corresponding method based on contextual information.
\citet{tao2024unveiling} introduce a sentence-level MGT detection method based on features from multiple levels and aspects. 
However, we carefully design a human-AI coauthored text generation pipeline different from the above. And we comprehensively reform and evaluate current detectors in both in-domain and out-of-domain settings, which is under-discussed in the literature.

\subsection{MGT Detectors Robustness}
The robustness of MGT detectors is a vital attribution for state-of-the-art detectors. 
\citet{shi2024red} first introduces the adversarial attack against MGT detectors, including word substitution and prompt attack to generate texts that deceive popular detectors.
\citet{wang2024stumbling} comprehensively study the detector's robustness towards a wide range of perturbation attack methods from character level to sentence level, finding significant performance degradation for most detectors under most attacks though only with limited access.
And \citet{dugan-etal-2024-raid} proposes RAID, a large-scale and challenging benchmark for testing MGT detector's robustness. 
Although these works conduct extensive robustness tests on MGT detectors and apply revision methods, \eg{} paraphrasing, as attacks, they overlook that fine-grained revision should be viewed as a coauthoring process.

\section{Task Definition}
\label{sec:data_task}




In general, we can categorize MGT detection tasks into the following three types of tasks based on different scales of authorship attribution:

\fakeparagraph{Document-level MGT detection} classifies the author attribution of text at the document level, \ie{} all text in the entire document will be labeled as machine-generated or human-written. Formally, given an input passage $\mathcal{T}$ and the detector $D_d$, we get the label of the entire document $\ell$. So the detection task can be represented as 
$D_d(\mathcal{T})=\ell$.

\fakeparagraph{Sentence-level MGT detection} detailed classifies each sentences. Some documents contain a mix of machine-generated and human-authored content that cannot be labeled as a whole. These contents usually contain a few sentences. So it is more reasonable to identify author attribution and label the entire document at the sentence level. The input passage $\mathcal{T}$ will be divided into sentence sequences $T_s=[s_1, s_2,..., s_n]$, and the label of the entire document will be $L_s=[\ell_1, \ell_2,..., \ell_n]$. The detection task can be defined as
$D_s(T_s)=L_s$.

\fakeparagraph{Word-level MGT detection} offers a more fine-grained text analysis. It could be hard to define and determine the text authorship even at the sentence level in some scenarios, \eg{} LLM paraphrasing on human-written sentences \citep{tripto2023ship}. Hence, detecting modifications\footnote{The definition of text modification and the threshold of its strength to shift authorship can be subjective, especially in different tasks and scenarios. In this paper, we do not aim to have a holistic claim or consensus on this issue, but we are trying to suggest one of the reasonable settings and practical pipelines to initially study the possibility and performance of word-level MGT detection methods.} of text at the word level may be meaningful. The input passage $\mathcal{T}$ will be divided into word sequences\footnote{We are using word level as the granularity of the task instead of token level to prevent the different split of different tokenizers. However, most methods output prediction on the token level. For multiple token words, we will do ensembling in the first place.} $T_w=[w_1, w_2,..., w_m]$ and labeling word-wise $L_w=[\ell_1, \ell_2,..., \ell_m]$. The detection task can be defined as:
$D_w(T_w)=L_w$.

\begin{figure*}[t]
\centering
  \includegraphics[width=0.75\textwidth]{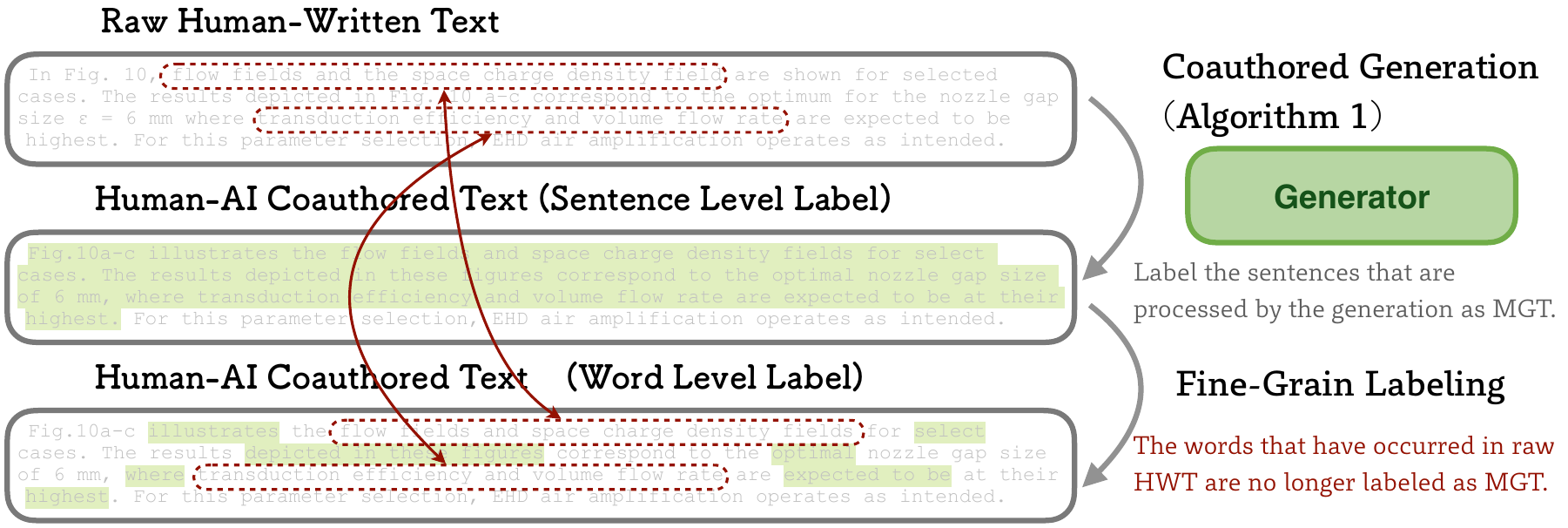}
  \caption{Grounded attribution labeling process on the word level and the sentence level setting in data construction. The fine-grain labeling process relies on word-span matching (in the red box and linked by curved arrows) between the hybrid texts before and after each revision turn.}
  \label{fig:transfer}
\end{figure*}

As we target fine-grained MGT detection, we choose to work on sentence-level and word-level detection tasks in this paper. 
The study pipeline overview is shown in \figref{fig:pip_study}.
To keep consistency between sentence-level and word-level prediction outputs, we formalize the task as sequence labeling at the word level first.
Basically, our task is to binary classify each word's authorship into HWT or MGT.
We input the entire text $\mathcal{T}$ and output sequence of word labels $L=[\ell_1, \ell_2,..., \ell_m]$. 
For sentence-level detection, we use the ensemble method to convert word-level labels to sentence-level labels. Specially, for each sentence $s_i$ composed of words $ [w_{i,1}, w_{i,2},..., w_{i,m}]$, we do majority voting as

\begin{align*}
    L_{s_i} &= \arg \max_c V_c(s_i) \\
    &= \arg \max_c \sum_{k=1}^{m} \mathbb{I}(L_{w_{i,k}} = c). 
\end{align*}


\begin{table*}[t]
\centering

\scalebox{0.8}{
\begin{tabular}{@{}cccccc@{}}
\toprule
\textbf{Domain}    & \textbf{Resources}    &  \textbf{Num} &  \textbf{\#MGT Frag.} & \textbf{MGT Portion} & \textbf{Length \#Word} \\ \midrule
News             & \texttt{Xsum}           &2,800     & 2.44        & 45\%              & 1,489          \\
Story          & \texttt{Wrting\_prompts} &2,800      & 3.97        & 40\%              & 1,964          \\
Scientific Paper & \texttt{Dagpap24}       &2,800      & 4.56       & 32\%              & 4,558          \\
Wikipedia        & \texttt{Wikipedia\_en}  &2,800      & 6.01        & 26\%              & 7,724          \\ \midrule
All        &                    -           &11,200      & 4.25         & 36\%
           & 3,934               \\ \bottomrule
\end{tabular}
}
\caption{The statistics of \dataset{} and its each domain. `{\#MGT Frag.}' refers to the number of LLM-authored text fragments, \ie{} the number of the LLM revision turns; and `{MGT Portion}' means the average portion of LLM-authored text fragments in each text in terms of word numbers. }
\label{tab:inf_dat}
\end{table*}

\section{Dataset Creation\label{sec:data}}

In this section, we present the details about our fine-grained MGT detection dataset, \dataset{}.
\figref{fig:transfer} illustrates the construction pipeline. 

\subsection{Raw HWT Resources}

\dataset{} comprises 11,200 human-authored texts from four different domains, including news articles from XSum \citep{narayan2018don}, story writing from WritingPrompts \citep{fan2018hierarchical}, scientific papers from Dagpap24 \citep{chamezopoulos-etal-2024-overview}, and Wikipedia contexts from Wikipedia\_en \citep{wikidump}.
We utilize these texts as initial human-authored manuscripts to construct human-AI coauthored content. 

For preprocessing, we filter out entries that are too short or of low quality. The filtering process is detailed in \appref{sec:appendix_check}.
We aim for an average length of around 4,000 words because it allows more turns of AI involvement. 
Moreover, it allows us to control the difficulty of tasks. 
\citet{solaiman2019release, he2023mgtbench} show that long text is less difficult to detect in document-level binary classification. 
Empirically, length of around 4,000 words alone performance differences between detectors are distinguishable.

\subsection{Construction Pipeline\label{sec:data_const}}




In order to make the generated dataset more diverse, we generated the text for the four domains on four instruction-tuning models, including Llama-3, Mixtral, GPT-4o mini, and GPT-4o. We provide details of these models' settings in Appendix \secref{sec:appendix_generator}. In addition, we apply different corresponding instructions for each model in the generation process. The instruction templates of each LLM for generation are shown in Appendix \secref{sec:appendix_instruction}.

In literature, the common dataset construction process of human-AI collaborative generation is feeding human-written prefixes to the model and then instructing the model to continue writing.
This process has two drawbacks: 
(\textit{i}) The annotation is given document-wise, either a turning point position or an overall label, instead of finer-grained. 
(\textit{ii}) It is completed in only a single turn without more interactions, which might oversimplify human-AI coauthoring.



We propose to construct a collaborative, word-level labeled dataset through multiple turns of revision interaction to deal with the above-mentioned issues.
Instead of continuing writing, we instruct LLMs to paraphrase specific parts of the contents of the raw HWTs, while extraction follows the sentence boundaries. 
We set the default temperature for LLM during paraphrasing.
We repeat this revision process multiple turns. The number of turns depends on the length of the initial raw texts.
Notably, we will locate and only revise the MGT spans in the hybrid texts to prevent ambiguity of authorship.
In detail, we apply NLTK toolkit \citep{bird2009natural} to divide the raw HWTs $\mathcal{T}$ into sequences of sentences $S = \mathcal{ST}(H) = [s_1, s_2,..., s_n ]$. Then we each turn select a continuous sentences span $S_{\text{span}} =[s_k , s_{k+1}, ..., s_{k+l}]$ as fragments based on the rules of Algorithm \ref{pseudo-code}. 
We query the instruct LLMs to paraphrase the selected span $S_{\text{span}}$ to get revised spans fragments $ S_{\text{span}}^{'} $, which is labeled as MGT in sentence level.
We replace $ S_{\text{span}} $ with its corresponding $ S_{\text{span}}^{'} $. Then follows the next turn.

For word-level labeling, we consider the lexical change in LLM revision turn. By default, all words in $ S_{\text{span}}^{'} $ is labeled as MGT. But if any word in $ S_{\text{span}}^{'} $ has occurred in the same tense in the corresponding $ S_{\text{span}} $, we keep its original attribution label.\footnote{We suggest that this is one of the many potential definitions of authorship transformation. The definition of attribution is not holistic and there can be different definitions and formulations in the fine-grained tasks. However, our intention is to have a checkable and reasonable heuristic. The discussion of the holistic definition of this issue is out of our scope.}
We illustrate the labeling process in \figref{fig:transfer}.

\begin{figure}[t]
\centering
\begin{subfigure}{0.235\textwidth}
    \centering
    \includegraphics[width=\textwidth]{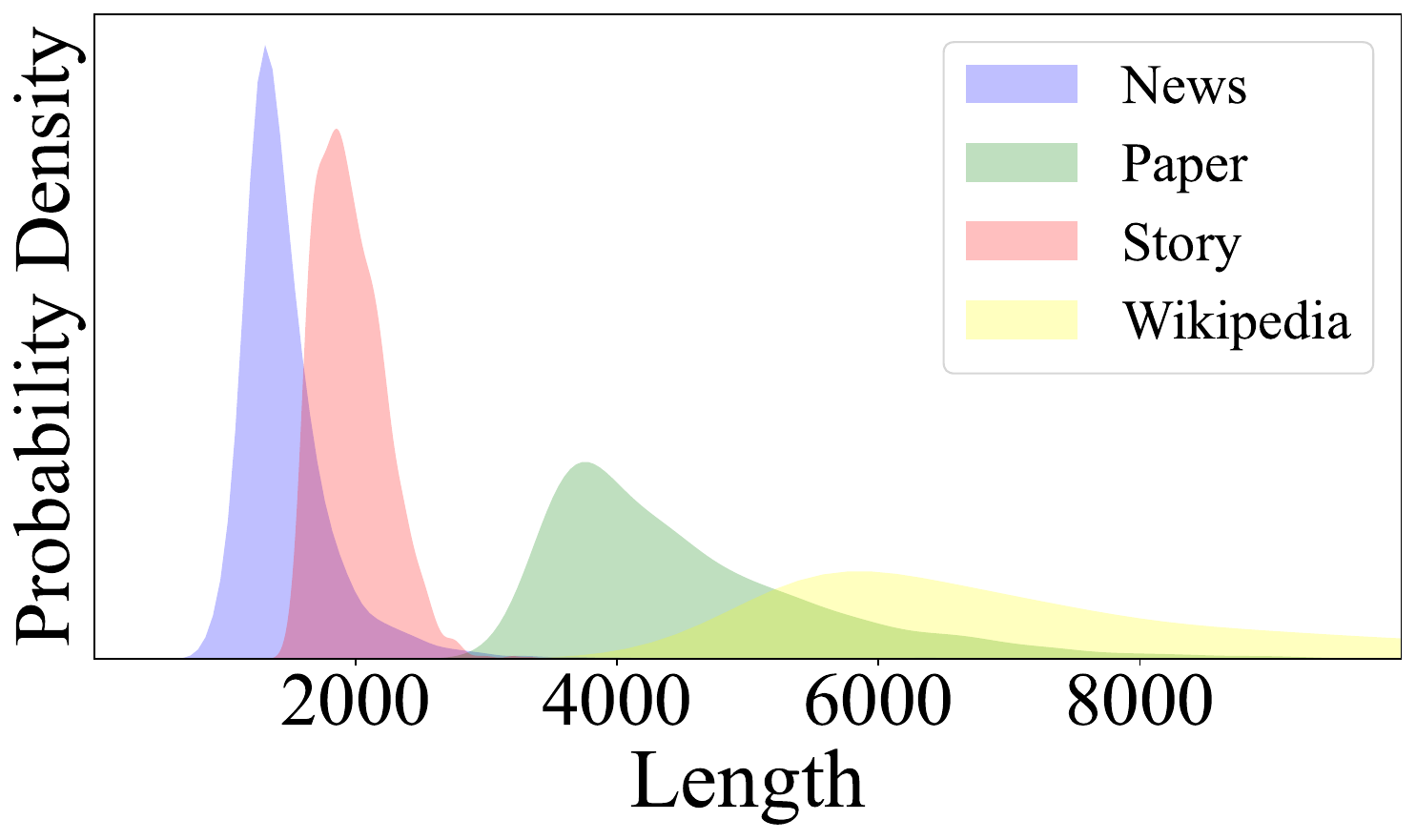}
\end{subfigure}
\begin{subfigure}{0.235\textwidth}
    \centering
    \includegraphics[width=\textwidth]{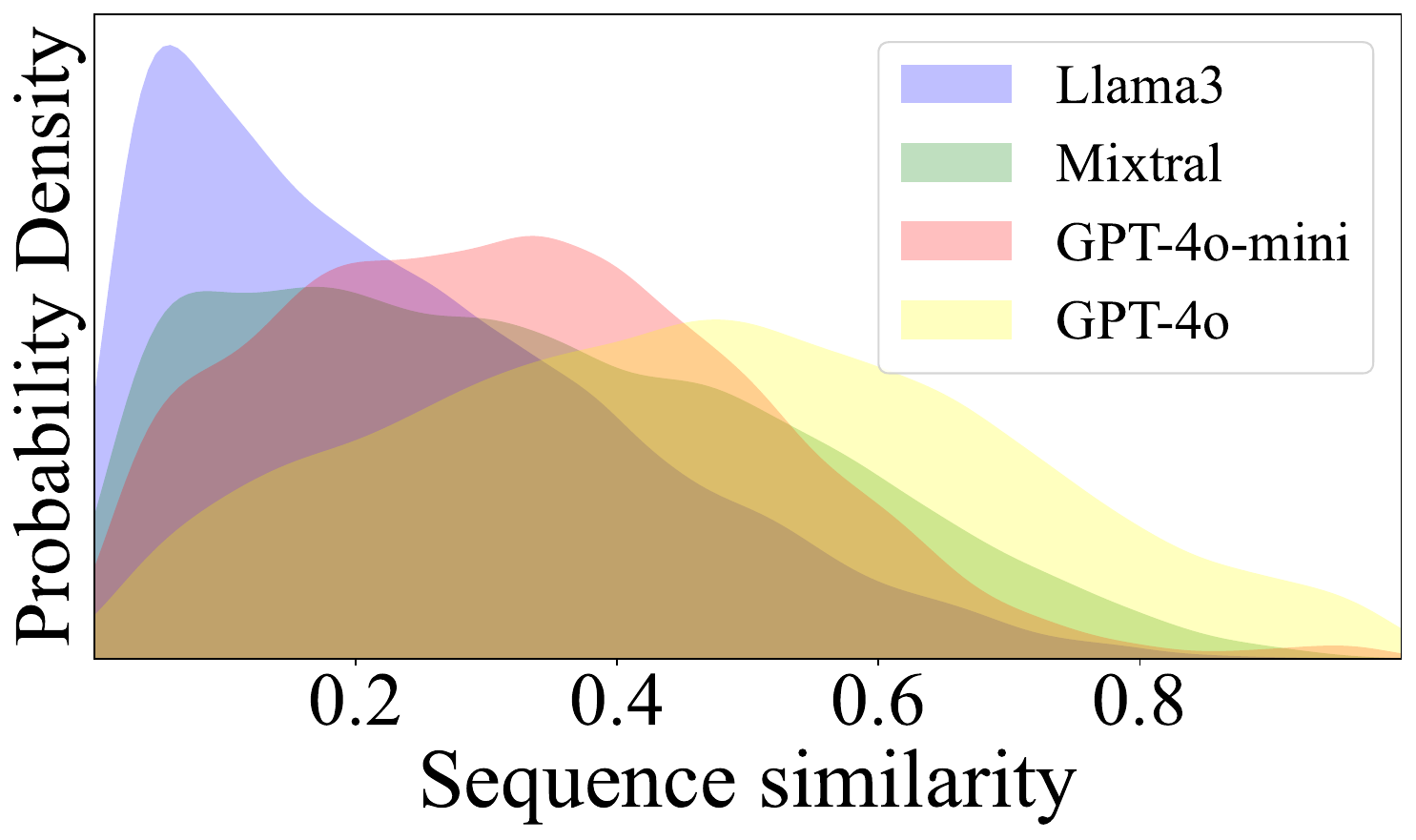}
\end{subfigure}
\caption{(Left) Length distribution of the texts from different domains. The paper and Wikipedia domains have much longer content than news and stories. (Right) Similarity of the texts from different generators. GPT-4o results in the most significant alterations while Llama-3 cases the least.}
\label{fig:length}
\end{figure}
 
\subsection{Dataset Analysis}
\tabref{tab:inf_dat} presents the statistics of the \dataset{} dataset.
To further sanity check the quality, we analyze length distribution and revision strength. 

\fakeparagraph{Length:}
While focusing on fine-grained detection of longer text, the length of texts still differs among different domains.
As shown in \figref{fig:length} Left, the orders of magnitude of word count ranged from 1,000 to 10,000 words. 
The paper and Wikipedia domains have much longer content than news and stories.

\fakeparagraph{Revision:}
We compute the sequence similarity between each pair of coauthored text and raw HWT. 
Specifically, we calculate the portion of substring shared with the coauthored text and its corresponding raw HWT.
A larger distance represents stronger revision by the LLM generators and might refer to a stronger signal in detection.
As shown in \figref{fig:length} Right, GPT-4o results in the most significant alterations while Llama-3 cases the least. 
This might refer to the generation capability of generators \citep{kirk2023understanding}.

\section{Experiment Setting}
\label{sec:exp_set}


To extensively evaluate the performance of different baselines on \dataset{}, we employ both in-domain and out-of-domain settings.

\fakeparagraph{In-domain setting.} We split the dataset into the training, validation, and test sets in a ratio of 4:1:2. For the finetune-based methods, we supervised fine-tuning them on the training set and report the performance on the test set.
For the metric-based methods, we pick the optimal threshold on the training set, which is the point on the ROC curve with the largest Youden index. We directly apply the threshold test on the test set.

\fakeparagraph{Out-of-distribution setting.} \dataset{} consists of four writing datasets from four different domains, and we use four different AI models as generators. Since overfitting on task and generator model is a common problem for detection methods, we want to evaluate whether the detection methods generalize well. We conducted out-of-distribution experiments in both these two dimensions. In the out-of-domain setting, we train on texts from one dataset (but mix texts from different generators) and test on text from other datasets. In the out-of-model setting, we will train on text generated from one model (mix datasets) and test on text generated from other generator models.



\fakeparagraph{Baselines.}
We evaluate three categories of baselines, including seven predominant detection methods, on \dataset{}. (\textit{\romannumeral 1}) {Finetune-based methods}, including SeqXGPT\footnote{We do not reform these methods because they are compatible with word-level MGT detection. \label{footnote:exp}} \citep{wang2023seqxgpt} and DeBERTa \citep{he2023debertav3}; 
(\textit{\romannumeral 2}) {Metric-based methods with perturbation}, including DetectGPT \citep{mitchell2023detectgpt}, Fast-DetectGPT \citep{bao2023fast}, and NPR in DetectLLM \citep{su2023detectllm}; (\textit{\romannumeral 3}) {Metric-based detectors without perturbation}, including GLTR\textsuperscript{\ref{footnote:exp}}\citep{gehrmann-etal-2019-gltr} and LRR in DetectLLM \citep{su2023detectllm}. We provide more details about baselines in Appendix \ref{sec:appendix_baseline} and their corresponding adaptions in Appendix \ref{sec:appendix_adaptions}.

\fakeparagraph{Metrics.} 
We employ macro f1-score, precision, recall, and AUC ROC as metrics to evaluate baselines. We present macro f1-score and AUC ROC, which are more suitable for imbalanced datasets, in the main text and other metrics in Appendix \ref{sec_appendix_pr}.






\section{Results\label{sec:result}}


\subsection{In-Domain Detection\label{word-ind}}
\fakeparagraph{Word-Level Detection.}
As shown in \tabref{tab:exp_in}, all SOTA detectors fail on our word-level coauthored text detection setting, except for DeBERTa.
The F1 scores and AUC scores for metric-based methods are around random prediction, indicating they are unable to conduct word-level detection.
DeBERTa reaches 0.831 in terms of F1-W, standing out from all competitors.
We suggest that semantic representation might be vital for fine-grained detection. 
As DeBERTa is a supervised fine-tuned detector based on embedding, it is accessible to more semantic features than metrics.
However, there is still room for accuracy improvement. 


\fakeparagraph{Sentence-Level Detection.}
In \tabref{tab:exp_in}, DeBERTa shows outperformance than other methods.
SeqXGPT also works and performs well in relative. 
In addition, Fast-DetectGPT and LLR perform best among the metrics-based detectors, which is consistent with their better performance in document-level detection \citep{dugan-etal-2024-raid}. 
In contrast, the performance of most metric-based detectors is inferior to random guesses, which emphasizes that their capability on document-level detection cannot effectively generalize to sentence-level under our setting. 
We suggest the reason is that simply ensembling is unable to adapt metrics computed at the word level to sentence-level detection.
The result emphasizes even at sentence-level \dataset{} tasks, most detectors are still far from perfect.

\begin{table}[t]
\centering

\scalebox{0.78}{
\renewcommand{\arraystretch}{1.1}
\begin{tabular}{@{\hspace{1.5mm}}lcccc@{\hspace{1.5mm}}}
\toprule[2pt]
\textbf{Detector}  & F1-W      $\uparrow$                & AUC-W $\uparrow$     & F1-S $\uparrow$               & AUC-S    $\uparrow$              \\ \midrule[1pt]
Random        & 0.433                     &-$^*$  & 0.497                     & -$^*$     \\ \midrule[1pt]
\multicolumn{5}{l}{\textbf{Finetune-based Methods}}                                                       \\ \midrule[1pt]
DeBERTa       & \underline{\textbf{0.831}}                     &  -$^*$        & \underline{\textbf{0.966}}                    & -$^*$      \\
SeqXGPT     & 0.513                     & -$^*$    & 0.674                     & -$^*$     \\ \midrule[1pt]
\multicolumn{5}{l}{\textbf{Metric-based Methods (w/ Perturb)}}                                                      \\ \midrule[1pt]
DetectGPT    & 0.375                     & 0.482  & 0.459               & 0.501        \\
NPR    & 0.414                     & 0.485         & 0.473               & 0.509  \\
Fast-DetectGPT   & \underline{0.501}                     & \underline{0.507}    & \underline{0.533}               & \underline{0.510}        \\ \midrule[1pt]
\multicolumn{5}{l}{\textbf{Metric-based Methods (w/o Perturb)}}                                                    \\ \midrule[1pt]
log prob    & \underline{0.479}                     & 0.482       & 0.444               & \underline{\textbf{0.511}}     \\
rank    & 0.441                     & 0.486       & 0.465               & 0.510      \\
log rank  & 0.439                     & 0.487    & 0.506               & \underline{\textbf{0.511}}        \\
entropy   & \underline{0.479}                     & \underline{\textbf{0.488}}      & 0.392               & \underline{\textbf{0.511}}       \\
LRR    & 0.475                     & 0.483        & \underline{0.516}               & 0.510  
\\ \bottomrule[2pt]
\end{tabular}
}
\caption{Results of IND fine-grained MGT detection. `-W' means at word level and `-S' means at sentence level. Asterisk (*) denotes that the AUC ROC is inaccessible since the method directly predicts without any threshold. Bold means the overall best performance, and underline means the best performance in the categories. `Random' refers to the result of random prediction. }
\label{tab:exp_in}
\end{table}
\begin{table*}[t]
\centering
\scalebox{0.55}{
\begin{tabular}{@{\hspace{2mm}}lcccccccccccccccc@{\hspace{2mm}}}
\toprule[2pt]
\multicolumn{1}{c|}{\multirow{2}{*}{\textbf{Detector}}} & \multicolumn{4}{c|}{News}                                                        & \multicolumn{4}{c|}{Paper}                                                       & \multicolumn{4}{c|}{Story}                                                       & \multicolumn{4}{c}{Wikipedia}                               \\
\multicolumn{1}{c|}{}                         & F1-W. & F1-S. & AUC-W.   & \multicolumn{1}{c|}{AUC-S.}       & F1-W. & F1-S.   & AUC-W.    & \multicolumn{1}{c|}{AUC-S.}               & F1-W. & F1-S.  & AUC-W.               & \multicolumn{1}{c|}{AUC-S.}                & F1-W. & F1-S. & AUC-W.               & AUC-S.       \\ \midrule[1pt]
random     & 0.465                          & 0.492                                     & -$^*$             & -$^*$          & 0.430                          & 0.498                             & -$^*$                          & -$^*$          & 0.465           & 0.496                         & -$^*$             & -$^*$         & 0.417                          & 0.495           & -$^*$             & -$^*$             \\ \midrule[1pt]
\multicolumn{17}{l}{\textbf{Fine-tune based Methods}}   \\ \midrule[1pt]
DeBERTa  & \underline{\textbf{0.826}}$\downarrow$ & \underline{\textbf{0.964}}$\downarrow$ & -$^*$                          & -$^*$                          & \underline{\textbf{0.776}}$\downarrow$              & \underline{\textbf{0.920}}$\downarrow$ & -$^*$                          & -$^*$                       & \underline{\textbf{0.830}}$\downarrow$ & \underline{\textbf{0.904}}$\downarrow$ & -$^*$           & -$^*$                       & \underline{\textbf{0.800}}$\downarrow$ & \underline{\textbf{0.935}}$\downarrow$ & -$^*$                          & -$^*$           \\
SeqXGPT  & 0.450$\downarrow$ & 0.508$\downarrow$ & -$^*$                          & -$^*$                          & 0.483$\downarrow$              & 0.458$\downarrow$ & -$^*$                          & -$^*$                       & 0.451$\downarrow$ & 0.573$\downarrow$ & -$^*$           & -$^*$                       & 0.490$\downarrow$ & 0.483$\downarrow$ & -$^*$                          & -$^*$  \\ \midrule[1pt]
\multicolumn{17}{l}{\textbf{Metric-based Methods(w)}}  \\ \midrule[1pt]
DetectGPT & 0.401$\uparrow$   & 0.403$\downarrow$ & 0.470$\downarrow$              & 0.494$\downarrow$              & 0.357$\downarrow$              & 0.441$\downarrow$ & 0.474$\downarrow$              & 0.496$\downarrow$           & 0.448$\uparrow$   & 0.434$\downarrow$ & 0.502$\uparrow$ & 0.509$\uparrow$             & 0.358$\downarrow$ & 0.434$\downarrow$ & 0.470$\downarrow$              & 0.496$\downarrow$ \\
NPR   & 0.444$\uparrow$   & 0.521$\uparrow$   & 0.483$\downarrow$              & \underline{0.512}$\uparrow$                & 0.403$\downarrow$              & 0.483$\uparrow$   & 0.480$\downarrow$              & 0.507$\downarrow$           & 0.476$\uparrow$   & \underline{0.559}$\uparrow$   & \underline{0.514}$\uparrow$ & \underline{0.529}$\uparrow$             & 0.403$\downarrow$ & \underline{0.445}$\downarrow$ & 0.483$\downarrow$              & \underline{0.510}$\uparrow$ \\
Fast-DetectGPT  & \underline{0.490}$\downarrow$ & \underline{0.544}$\uparrow$   & \underline{0.494}$\downarrow$              & 0.510 \text{-} & \underline{0.467}$\downarrow$              & \underline{0.532}$\downarrow$ & \underline{\textbf{0.494}}$\downarrow$              & \underline{\textbf{0.509}}$\downarrow$           & \underline{0.491}$\downarrow$ & 0.552$\uparrow$   & 0.510$\uparrow$ & 0.517$\uparrow$             & \underline{0.459}$\downarrow$ & 0.432$\downarrow$ & \underline{\textbf{0.496}}$\downarrow$              & 0.508$\downarrow$ \\ \midrule[1pt]
\multicolumn{17}{l}{\textbf{Metric-based Methods(w/o)}}  \\ \midrule[1pt]
log prob  & \underline{0.485}$\uparrow$   & 0.509$\uparrow$   & 0.483$\uparrow$                & 0.517$\uparrow$                & 0.473$\downarrow$              & 0.466$\uparrow$   & 0.480$\downarrow$              & \underline{\textbf{0.509}}$\downarrow$           & 0.505$\uparrow$   & 0.560$\uparrow$   & 0.520$\uparrow$ & 0.537$\uparrow$             & 0.467$\downarrow$ & 0.356$\downarrow$ & 0.481$\downarrow$              & \underline{\textbf{0.513}}$\uparrow$ \\
rank  & 0.467$\uparrow$   & 0.508$\uparrow$   & 0.484$\downarrow$              & 0.514$\uparrow$                & 0.435$\downarrow$              & 0.507$\uparrow$   & 0.483$\downarrow$              & 0.508$\downarrow$           & 0.493$\uparrow$   & 0.563$\uparrow$   & 0.513$\uparrow$ & 0.530$\uparrow$             & 0.430$\downarrow$ & 0.412$\downarrow$ & 0.485$\downarrow$              & 0.511$\uparrow$ \\
log rank & 0.467$\uparrow$   & 0.543$\uparrow$   & 0.487\text{-} & 0.516$\uparrow$                & 0.433$\downarrow$              & 0.497$\downarrow$ & 0.484$\downarrow$              & \underline{\textbf{0.509}}$\downarrow$           & 0.493$\uparrow$   & 0.534$\uparrow$   & 0.517$\uparrow$ & 0.532$\uparrow$             & 0.429$\downarrow$ & 0.337$\downarrow$ & \underline{0.486}$\downarrow$              & 0.512$\uparrow$ \\
entropy  & 0.484$\uparrow$   & 0.467$\uparrow$   & \underline{\textbf{0.499}}$\uparrow$                & \underline{\textbf{0.525}}$\uparrow$                & \underline{0.478}$\downarrow$              & 0.397$\uparrow$   & \underline{0.488}\text{-} & \underline{\textbf{0.509}}$\downarrow$           & 0.499$\uparrow$   & 0.516$\uparrow$   & \underline{\textbf{0.531}}$\uparrow$ & \underline{\textbf{0.543}}$\uparrow$             & \underline{0.472}$\downarrow$ & 0.328$\downarrow$ & 0.485$\downarrow$              & 0.512$\uparrow$  \\
LRR & 0.484$\uparrow$   & \underline{0.551}$\uparrow$   & 0.486$\uparrow$                & 0.515$\uparrow$                & 0.475\text{-} & \underline{0.509}$\downarrow$ & 0.480$\downarrow$              & 0.508$\downarrow$           & \underline{0.507}$\uparrow$   & \underline{0.572}$\uparrow$   & 0.521$\uparrow$ & 0.534$\uparrow$             & 0.464$\downarrow$ & \underline{0.482}$\downarrow$ & 0.483\text{-} & \underline{\textbf{0.513}}$\uparrow$ \\ \bottomrule[2pt]
\end{tabular}
}
\caption{Results of out-of-domain fine-grained MGT detection. Detectors are tested on the mentioned domain and trained on the rest. ‘-W’ means
at word level and ‘-S’ means at sentence level. Asterisk (*) denotes that the AUC ROC is inaccessible since the method directly predicts without any threshold. Bold means the overall best performance, and underline means the best performance in the categories. `Random' refers to the result of random prediction.  $\uparrow$ indicates the method has a better performance than in-domain, $\downarrow$ indicates the method has a worse performance than in-domain.}
\label{tab:exp_out_domain}
\end{table*}
\begin{table*}[t]
\centering
\scalebox{0.55}{
\begin{tabular}{@{\hspace{2mm}}lcccccccccccccccc@{\hspace{2mm}}}
\toprule[2pt]
\multicolumn{1}{c|}{\multirow{2}{*}{\textbf{Detector}}} & \multicolumn{4}{c|}{Llama3}                                                      & \multicolumn{4}{c|}{Mixtral}                                                     & \multicolumn{4}{c|}{GPT-4o}                                                      & \multicolumn{4}{c}{GPT-4o-mini}                             \\
\multicolumn{1}{c|}{}                         & F1-W. & F1-S.  & AUC-W.               & \multicolumn{1}{c|}{AUC-S.}             & F1-W. & F1-S.  & AUC-W.               & \multicolumn{1}{c|}{AUC-S.}              & F1-W. & F1-S.  & AUC-W.               & \multicolumn{1}{c|}{AUC-S.}               & F1-W. & F1-S. & AUC-W.               & \multicolumn{1}{c}{AUC-S.}                \\ \midrule[1pt]
random      & 0.449             & 0.495             & -$^*$                          & -$^*$                          & 0.434             & 0.497             & -$^*$             & -$^*$                      & 0.411             & 0.502             & -$^*$                          & -$^*$                          & 0.435             & 0.500             & -$^*$             & -$^*$  \\ \midrule[1pt]
\multicolumn{17}{l}{\textbf{Fine-tune based Methods}} \\ \midrule[1pt]
DeBERTa   & \underline{\textbf{0.854}}$\uparrow$   & \underline{\textbf{0.976}}$\uparrow$   & -$^*$                          & -$^*$                          & \underline{\textbf{0.808}}$\downarrow$ & \underline{\textbf{0.951}}$\downarrow$ & -$^*$             & -$^*$                      & \underline{\textbf{0.768}}$\downarrow$ & \underline{\textbf{0.932}}$\downarrow$ & -$^*$                          & -$^*$                          & \underline{\textbf{0.815}}$\downarrow$ & \underline{\textbf{0.968}}$\uparrow$   & -$^*$             & -$^*$  \\
SeqXGPT & 0.462$\downarrow$ & 0.653$\downarrow$ & -$^*$                          & -$^*$                          & 0.492$\downarrow$ & 0.666$\downarrow$ & -$^*$             & -$^*$                      & 0.498$\downarrow$ & 0.654$\downarrow$ & -$^*$                          & -$^*$                          & 0.484$\downarrow$ & 0.661$\downarrow$ & -$^*$             & -$^*$ \\ \midrule[1pt]
\multicolumn{17}{l}{\textbf{Metric-based Methods(w)}} \\ \midrule[1pt]
DetectGPT  & 0.397$\uparrow$   & 0.454$\downarrow$ & 0.491$\uparrow$                & 0.501\text{-} & 0.382$\uparrow$  & 0.460$\uparrow$   & 0.486$\uparrow$   & 0.503$\uparrow$            & 0.347$\downarrow$ & 0.470$\uparrow$   & 0.473$\downarrow$              & 0.504$\uparrow$                & 0.374$\downarrow$ & 0.451$\downarrow$ & 0.471$\downarrow$ & 0.498$\downarrow$ \\
NPR & \underline{0.439}$\uparrow$   & 0.476$\uparrow$   & 0.499$\uparrow$                & 0.513$\uparrow$                & 0.425$\uparrow$   & 0.481$\uparrow$   & 0.493$\uparrow$   & \underline{0.512}$\uparrow$            & 0.388$\downarrow$ & 0.497$\uparrow$   & 0.470$\downarrow$              & \underline{0.507}$\downarrow$              & 0.406$\downarrow$ & 0.481$\uparrow$   & 0.467$\downarrow$ & \underline{0.506}$\downarrow$ \\
Fast-DetectGPT  & 0.393$\downarrow$ & \underline{0.553}$\uparrow$   & \underline{\textbf{0.517}}$\uparrow$          & \underline{\textbf{0.518}}$\uparrow$       & \underline{0.473}$\downarrow$ & \underline{0.528}$\downarrow$ & \underline{\textbf{0.506}}$\downarrow$ & \underline{0.512}$\uparrow$            & \underline{0.447}$\downarrow$ & \underline{0.520}$\downarrow$ & \underline{0.480}$\downarrow$              & 0.504$\downarrow$              & \underline{0.454}$\downarrow$ & \underline{0.519}$\downarrow$ & \underline{0.478}$\downarrow$ & 0.504$\downarrow$ \\ \midrule[1pt]
\multicolumn{17}{l}{\textbf{Metric-based Methods(w/o)}}  \\ \midrule[1pt]
log prob  & 0.476$\downarrow$ & 0.459$\uparrow$   & 0.496$\uparrow$                & 0.514$\uparrow$                & 0.474$\downarrow$ & 0.451$\uparrow$   & 0.490$\uparrow$   & \underline{\textbf{0.514}}$\uparrow$            & \underline{0.477}$\downarrow$ & 0.434$\downarrow$ & 0.468$\downarrow$              & 0.507$\downarrow$              & \underline{0.481}$\uparrow$   & 0.456$\uparrow$   & 0.465$\downarrow$ & 0.508$\downarrow$ \\
rank  & 0.461$\uparrow$   & 0.468$\uparrow$   & 0.501$\uparrow$                & 0.514$\uparrow$                & 0.450$\uparrow$   & 0.473$\uparrow$   & \underline{0.496}$\uparrow$   & \underline{\textbf{0.514}}$\uparrow$            & 0.417$\downarrow$ & 0.455$\downarrow$ & 0.471$\downarrow$              & 0.506$\downarrow$              & 0.435$\downarrow$ & 0.463$\downarrow$ & 0.468$\downarrow$ & 0.506$\downarrow$ \\
log rank & 0.459$\uparrow$   & 0.406$\downarrow$ & \underline{0.502}$\uparrow$                & \underline{0.516}$\uparrow$                & 0.444$\uparrow$   & 0.413$\downarrow$ & \underline{0.496}$\uparrow$   & \underline{\textbf{0.514}}$\uparrow$            & 0.416$\downarrow$ & 0.495$\downarrow$ & 0.472$\downarrow$              & 0.507$\downarrow$              & 0.433$\downarrow$ & 0.500$\downarrow$ & 0.469$\downarrow$ & 0.507$\downarrow$ \\
entropy & \underline{0.483}$\uparrow$   & 0.377$\downarrow$ & 0.488\text{-} & 0.508$\downarrow$              & \underline{0.482}$\uparrow$   & 0.394$\uparrow$   & 0.489$\uparrow$   & 0.512$\uparrow$            & 0.471$\downarrow$ & 0.383$\downarrow$ & \underline{\textbf{0.488}}\text{-} & \underline{\textbf{0.511}}\text{-} & 0.478$\downarrow$ & 0.397$\uparrow$   & \underline{\textbf{0.487}}$\downarrow$ & \underline{\textbf{0.513}}$\uparrow$ \\
LRR & 0.481$\uparrow$   & \underline{0.529}$\uparrow$   & 0.494$\uparrow$                & 0.512$\uparrow$                & 0.476$\uparrow$   & \underline{0.521}$\uparrow$   & 0.488$\uparrow$   & 0.511$\uparrow$            & 0.466$\downarrow$ & \underline{0.507}$\downarrow$ & 0.472$\downarrow$              & 0.507$\downarrow$              & 0.473$\downarrow$ & \underline{0.512}$\downarrow$ & 0.472$\downarrow$ & 0.508$\downarrow$ \\  \bottomrule[2pt]
\end{tabular}
}

\caption{Results of out-of-model fine-grained MGT detection. Detectors are tested on the mentioned domain and trained on the rest. ‘-W’ means
at word level and ‘-S’ means at sentence level. Asterisk (*) denotes that the AUC ROC is inaccessible since the method directly predicts without any threshold. Bold means the overall best performance, and underline means the best performance in the categories. `Random' refers to the result of random prediction. $\uparrow$ indicates the method has a better performance than in-domain, $\downarrow$ indicates the method has a worse performance than in-domain.}
\label{tab:exp_out_model}
\end{table*}

\begin{table*}[t]
\scalebox{0.75}{
\begin{tabular}{@{\hspace{2mm}}ccccccccccc@{\hspace{2mm}}}
\toprule
Method  & DeBERTa & \multicolumn{1}{c|}{SeqXGPT}   & DetectGPT & NPR  & \multicolumn{1}{c|}{Fast-DetectGPT} & logprob   & rank   & logrank   & entropy   & LRR     \\ \midrule
error-S & \underline{\textbf{1.78\%}}  & 12.00\% & \underline{10.70\%}   &  21.01\% & 14.75\%   & 23.70\%  & 21.92\%  & 18.32\%  & 29.10\% & \underline{17.44\%} \\
error-W & \underline{\textbf{1.84\%}}  & 10.00\% & 43.48\%   & 32.37\% & \underline{12.65\%}  & \underline{10.86\%} & 24.64\% & 24.96\% & 10.89\% & 12.00\% \\ \bottomrule
\end{tabular}
}
\caption{Results of IND document level AI rate prediction. ‘-W’ means
the difference between the predicted ratio of AI labels and the true ratio of AI labels at word level, and ‘-S’ denotes at sentence level. Bold means the overall best performance, and underlined means the best performance in the categories.}
\label{tab:exp_ai_rate}
\end{table*}

\subsection{Out-of-Distribution Detection\label{ood}}
Previous methods often show limitations on out-of-distribution (OOD) datasets \citep{mitchell2023detectgpt}. The performance of detectors will decrease due to variations in prompts \citep{koike2023you}, sampling methods, and the inherent differences in length, style, and quality among texts \citep{he2023mgtbench}. So we intend to test the performance of each method on the out-of-domain and out-of-model sentence-level detection dataset.

The results of them are shown in \tabref{tab:exp_out_domain} and \tabref{tab:exp_out_model}.
Overall, metric-based methods without perturbation have the best generalizability, and finetune-based methods are the worst. 
In addition, we find an interesting phenomenon that almost all methods achieve better OOD performance than IND when trained on paper and Wikipedia domains.
We propose that it may be because texts from these domains are longer (shown in \figref{fig:length} Left) and more diverse in language pattern, causing it to be easier to generalize from them to other more consistent domains. 
Similarly, most methods achieve better OOD performance when trained on the GPT-4o families corpus than IND.
We suggest that it may be easier for detectors to generalize from paraphrasers which prefer to make more significant modifications (shown in \figref{fig:length} Right).

\begin{figure}[ht]
  \includegraphics[width=0.95\columnwidth]{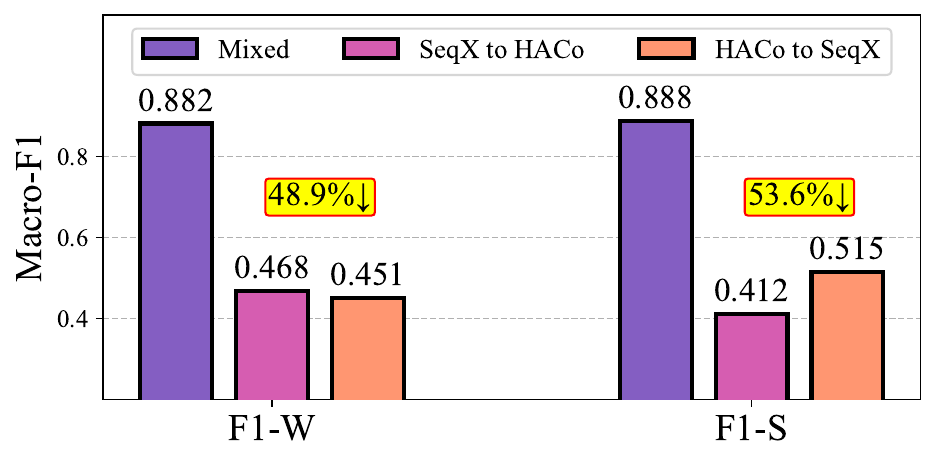}
  \caption{F1 score of revision mode generalization for the word- and sentence-level detection (abbreviated as `-W' and `-S'). For reference, `Mixed' is the performance that trains and tests detectors on a mixed set of the two datasets. Results show that DeBERTa suffers from generalizing between different revision modes.}
  \label{fig:settings}
\end{figure}

\subsection{Document Level AI Rate Prediction}
The AI ratio—defined as the percentage of machine-generated content in a document—is a valuable metric for assessing the degree of AI involvement versus human contribution. Although most models struggle to detect AI-generated content with fine-grained precision, they may still provide a reasonable estimate of the overall AI ratio, which is a coarser but still informative measure. Notably, this remains a more challenging task than the document-level sequence labeling task addressed by \citet{wang2023seqxgpt}.

In this experiment, we compute the prediction error ratio, which is the absolute difference between the predicted and true AI label ratios for each document. A lower error percentage indicates more accurate document-level predictions. These predictions are derived by aggregating sentence-level or word-level labels. The results, presented in \tabref{tab:exp_ai_rate}, show that the fine-tuned method consistently outperforms others. However, certain approaches—such as DetectGPT, Fast-DetectGPT, and LRR—also achieve relatively low prediction errors, demonstrating promising potential.


\section{Analysis and Findings}
\label{sec:ana}


\subsection{Limitation of the Current SOTA}
DeBERTa, a supervised method, achieves the best performance in all settings (\secref{sec:result}). Though supervised methods are often criticized due to their weak generalization capability, in our experiments, DeBERTa does not show a significantly poor performance in OOD experiments. We further analyze if the supervised method has other shortcomings. We take DeBERTa as the analyzed backbone pre-trained model since it is the SOTA across others, including RoBERTa \citep{liu2019roberta}, XLNet \citep{yang2019xlnet}, and ELECTRA \citep{ClarkLLM20}. Results are shown in \appref{sec_appendix_backbone}.

As \citet{gao2024llm} find current detectors struggle to generalize across different revised operations. We evaluate the generalizability of DeBERTa across \dataset{} and SeqXGPT-Bench \citep{wang2023seqxgpt}, which has a different revision pipeline.
We evaluate both sentence-level detection and word-level detection.
The results are shown in \figref{fig:settings}. 
It is still hard for DeBERTa to generalize between different revision modes, as the gap between the two datasets is greater and more complex compared with the previous domain and generator generalization. 

\subsection{Zero-Shot Detection}
\label{sec:als_zero}

\begin{table}[]
\centering
\renewcommand{\arraystretch}{1.1}
\scalebox{0.9}{
\begin{tabular}{@{\hspace{1.5mm}}cll@{\hspace{1.5mm}}}
\toprule[1.5pt]
\textbf{Setting}                   & \multicolumn{1}{c}{F1-W.} & \multicolumn{1}{c}{F1-S.} \\ \midrule[1pt]
In-domain                 & \multicolumn{1}{c}{0.571} & \multicolumn{1}{c}{0.597} \\ \midrule[1pt]
Out-of-Domain (News)       & \multicolumn{1}{c}{0.530} & \multicolumn{1}{c}{0.514} \\
Out-of-Domain (Paper)      & 0.543                     & 0.572                     \\
Out-of-Domain (Story)      & 0.550                     & 0.523                     \\
Out-of-Domain (Wikipedia)  & \underline{\textbf{0.578}}                     & \underline{\textbf{0.665}}
\\ \midrule[1pt]
Out-of-Model (Llama)       & 0.536                     & \underline{0.633}                     \\
Out-of-Model (Mixtral)     & 0.551                     & 0.584                     \\
Out-of-Model (GPT-4o)      & 0.547                     & 0.521                     \\
Out-of-Model (GPT-4o-mini) & \underline{0.564}                     & 0.598                     \\ \bottomrule[1.5pt]
\end{tabular}
}
\caption{Performance of the encoder-frozen DeBERTa. `-W' means at word level and `-S' means at sentence level. Bold means the overall best performance, and underline means the best performance in the categories.}
\label{tab:exp_frozen}
\end{table}

All the detectors we have evaluated in previous experiments were trained or optimized to some extent. 
However, in real scenarios, it is difficult for us to have a supervised phase before detecting unknown data distribution. 
Therefore, we explore the possibilities of fine-grained zero-shot detection. 

\fakeparagraph{Finetune-based method.} We first try a soft setting that freezes the encoding module of DeBERTa and sorely tunes the classification heads on the training set. IND and OOD results are shown in \tabref{tab:exp_frozen}.
The performance of frozen DeBERTa degrades significantly, which suggests it is far from able to reliably do zero-shot detection. 
A possible enhancement could be having a more complex process handling the semantic encoding or using data augment to learn a more general encoding module before applying it on zero-shot deployment. 

\fakeparagraph{Metric-based method.} 
Previous experiments show that, even optimized threshold on the IND training set, the performance is far from usable.
As a softer setting, we try to directly use sentence-level metrics to do the sentence-level detection task. The results are shown in \appref{sec_appendix_sentence}. However, the detectors still struggle, as the performance is close to majority voting from the word level. 
We suggest these metrics might have different distributions on different granularity. And they are more indistinguishable in lower granularity.
Hence, it is less beneficial for applying metrics from a larger granularity for predicting labels of the smaller granularity, as attempted by SeqXGPT and \citet{zhang-etal-2024-machine}. 
We suggest that a patch might be fine-tuning the base model for metrics computing to adapt it to conditional probability distributions of lower-granularity tasks.

\begin{figure}[t]
  \includegraphics[width=0.95\columnwidth]{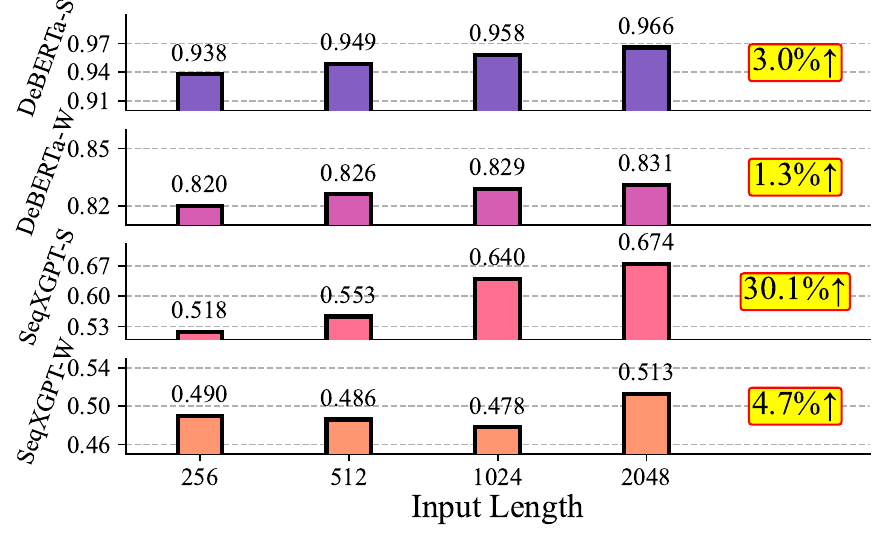}
  \caption{F1 score of DeBERTa and SeqXGPT with different lengths of input chunk for the word- and sentence-level detection (abbreviated as `-W' and `-S'). Generally, a larger length leads to better performance.}
  \vspace{-15pt}
  \label{fig:input_length}
\end{figure}

\subsection{Influence of Context Window}
Moreover, we suggest that the context window of the classification model is a bottleneck when detecting long coauthored texts.
LLM will truncate texts after a maximum input length limitation, but the length of the texts in \dataset{} sometimes exceeds.
In practice, we need to do chunking on long texts for some detectors that only support smaller context windows.
However, we argue that chunking will harm the detection performance by reducing the context, especially for fine-grained settings.
Results in \figref{fig:input_length} support the argument, showing consistent performance improvement when DeBERTa and SeqXGPT increase the text length of the input chunk, both at the sentence level and the word level.



\section{Conclusion}
In this paper, we introduced \dataset{}, a novel task and benchmark for word-level and sentence-level fine-grained co-authored MGT detection. It is constructed by a pipeline simulating human-AI collaborative generation processes.
We reform the mainstream detectors and evaluate their performance in \dataset{}, in in-domain and out-of-domain settings.
The results indicate that it is still challenging for current detectors to finish the \dataset{} task. 
Following this, we do some analysis and additional experiments to gather the intuition of current limitations, influential factors, and future enhancement methods.

\section*{Ethical Consideration}
Our study demonstrates that it is a challenge for existing detection methods to get a satisfying fine-grained detection performance in \dataset{}. However, our study aims purely for scientific exploration and to promote the development in this field. We strongly oppose the misuse of the construction method of \dataset{} to evade detection, such as homework assignments and fake news generation.
Instead, MGT detection should be used as a tool for prevention, deterrence, and warning of misuse and potential harm.

We affirm that all open-source resources utilized in our study, including detectors, language models, and datasets, have been employed to comply with their original licensing agreements.

\section*{Limitation}
Our work has the following limitations:
\textit{i}) Although \dataset{} is already diverse in some aspects, it is not multilingual. Hence, we did not evaluate the performance of the baselines in other languages and out-of-language detection. 
\textit{ii}) In \dataset{}, to consider situations that are more likely to occur in reality, the human text is polished by only one model. More complex multi-LLM collaboration together with human-LLM interaction should be focused on in future studies such as multi-model. 
\textit{iii}) In our study, though we obtained some possible improvements applicable to the existing detectors for fine-grained detection via analysis, we do not further focus on proposing a specific outperforming detection method as a solution. 
\textit{iv}) A limited number of prompt designs in our coauthored text generation pipeline are explored.

\section*{Acknowledgement}
We would like to thank all reviewers for their insightful comments and suggestions to help improve the paper. This work was supported by the National Nature Science Foundation of China (No. 62192781, No. 62272374), the Natural Science Foundation of Shaanxi Province (2024JC-JCQN-62), the National Nature Science Foundation of China (No. 62202367, No. 62250009), the Key Research and Development Project in Shaanxi Province No. 2023GXLH-024, Project of
China Knowledge Center for Engineering Science and Technology, and Project of Chinese academy of engineering “The Online and Offline Mixed Educational Service System for ‘The Belt and Road’ Training in MOOC
China”, and the K. C. Wong Education Foundation.

\bibliography{custom}

\appendix

\section{Dataset Details}
\label{sec:appendix}
\subsection{Unqualified Case in Human Data Check}
\label{sec:appendix_check}
We used human checks in the study to filter out low-quality data when selecting raw human texts. We found unqualified data in all raw text resources, the major types of them are:

\begin{itemize}
\item {Sport game broadcasts that include too many names and team names}, the case is shown in \figref{fig:flt_case_1}.
\item {Stories containing comments and replies}, the case is shown in \figref{fig:flt_case_2}.
\item {Papers that contain too many formula sections}, the case is shown in \figref{fig:flt_case_3}.
\item {Wikipedia that contain a list}, the case is shown in \figref{fig:flt_case_4}.
\item {Text containing HTML tags}, the case is shown in \figref{fig:flt_case_5}.
\end{itemize}

The red highlights in the figures are the contents recognized under the unqualification type above.
\subsection{Detailed Generation Setting}
\label{sec:appendix_setting}
Considering that different generation strategies affect the performance of the detectors, to give diversity to the models we obtained, we used different hyperparameters on the different generation models according to their default setting, as shown in \figref{fig:hyper_gen}. In addition, since the original text may contain punctuation or coded formats that cannot properly be tokenized by the model (\eg{} URLs), as well as sentences that are irrelevant {\eg{} ``\texttt{Media playback is not supported on this device}'', we performed a pre-processing operation on it before generation. After generating the text, we also perform a post-processing operation on it, we discard the generated text that is smaller than the length of the original text, to avoid generating text that does not comply with the instructions.

\begin{algorithm*}[h]
\caption{Human-AI coauthored text generation\label{pseudo-code}\label{algo}}
  \begin{algorithmic}[1]
    \Require
      raw HWT $\mathcal{T}$;
      generative model $\mathcal{M}$ for revision; 
      sentence tokenizer $\mathcal{ST}$;
      preset bound of span sentence number $[l_{\text{min}}, l_{\text{max}}]$;
    \Ensure
      human-AI coauthored text $C$;
    \State Divide $H$ into sequences of sentences $S = \mathcal{ST}(H) = [s_1, s_2,..., s_n]$;
    \State Assign $P = [p_{s_1}, p_{s_2}, ..., p_{s_n}] $ where $p_{s_i}$ is the beginning position of $s_i$, $i \leq n, i \in \mathbb{Z}^+$;
    \State Random select a sentence boundary position $p_{s_k}$ where $ k \leq \lceil \frac{n}{2} \rceil, k \in \mathbb{Z}^+$;
    \While{$ k + l_{\text{min}} \leq n$}
    \State Random select span length $l \in \mathbb{Z}^+, l \in [l_{\text{min}}, min(n-k, l_\text{max})]$; 
    \State Sample continuous sentence spans $S_{\text{span}} =[s_k , s_{k+1}, ..., s_{k+l}]$; 
    \State Replace $ S_{\text{span}}$ with $ S_{\text{span}}^{'} = \mathcal{M}(S_{\text{span}}, \text{context = }S \setminus S_{\text{span}}) $;
    \State Update $S$ and $P$;
    \State Update $n, k$ if sentence number changes\footnotemark;
    \State Assign $k_{\text{end}} = k + len(S_{\text{span}}^{'}) $;
    \State Random select $k \in [ k_{\text{end}} , \lceil \frac{n+k_{\text{end}}}{2} \rceil ]$;
    
    \EndWhile \\
    \Return $C = joint(S)$;
  \end{algorithmic}
  \label{code:fram:select} 
\end{algorithm*}
\footnotetext{In few cases, after revision $len(S_{\text{span}}^{'}) \neq l$. If so, we update $k$ as new index of the sentence
 that ends the paraphrased fragment.}

\subsection{Generators}
\label{sec:appendix_generator}
\fakeparagraph{Llama-3-8B-Instruct} \citep{llama3modelcard} is a collection of pre-trained and instruction-tuned generative text models. The Llama 3 instruction-tuned models are optimized for dialogue use cases and outperform many of the available open-source chat models on common industry benchmarks.

\fakeparagraph{Mixtral-8x7B-Instruct-v0.1} \citep{mistral2023moe} is a sparse mixture-of-experts network. It is a decoder-only model where the feedforward block picks from a set of 8 distinct groups of parameters. At every layer, for every token, a router network chooses two of these groups to process the token and combine their output additively.

\fakeparagraph{GPT-4o} \citep{openai2024gpt4o} is an autoregressive omni model, which accepts as input any combination of text, audio, image, and video and generates any combination of text, audio, and image outputs. It matches GPT-4 Turbo performance on text in English and code, with significant improvement on text in non-English languages. The version of GPT4 we used is gpt-4o-2024-08-06.

\fakeparagraph{GPT-4o mini} \citep{openai2024gpt4omini} enables a broad range of tasks with its low cost and latency and scores 82\% on MMLU and currently outperforms GPT-4 on chat preferences in LMSYS leaderboard. The version of GPT4 we used is gpt-4o-mini-2024-07-18.

\begin{table*}[t]
\small
\centering
\renewcommand{\arraystretch}{1.1}
\begin{tabular}{c|cccc}
\toprule
Model         & \multicolumn{1}{c}{Llama3} & \multicolumn{1}{c}{Mixtral} & \multicolumn{1}{c}{GPT-4o} & GPT-4o-mini \\ \toprule
Min Sentence & \multicolumn{4}{c}{10}                                                                                \\ \midrule 
Max Sentence & \multicolumn{4}{c}{15}                                                                                \\  \midrule
Top P         & \multicolumn{4}{c}{0.9}                                                                               \\  \midrule
Temperature   & \multicolumn{2}{c|}{0.6}                                   & \multicolumn{2}{c}{0.8}                  \\  \midrule
Context Window    & \multicolumn{2}{c|}{2048}                                  & \multicolumn{2}{c}{1024} 
\\
\bottomrule
\end{tabular}
\caption{Hyperparameters of generation models. `Min Sentence' is the minimal number of sentences modified once time. `Max Sentence' is the maximum.}
\label{fig:hyper_gen}
\end{table*}

\subsection{Instructions for Generation}
\label{sec:appendix_instruction}
We generated the text for the four domains on four models, using different corresponding instructions in the generation process. To investigate the performance of existing detection methods, we would like to use the prompt that not only enables rewriting, but also evades detection. Previous works have used a number of different prompts to evade detection through rewriting, and we employ a simple one for this purpose. We show the instruction template of LLM’s operation in \figref{fig:ins_chat_tem}, \figref{fig:ins_llama_tem}, \figref{fig:ins_mixtral_tem}. The \texttt{<passage>} will be replaced with the original sentences that need to be polished and the \texttt{<role>} will be replaced with the role that corresponds to the domain of the text. news corresponding to \texttt{news writers}, story corresponding to \texttt{story writer}, paper corresponding to \texttt{scientific paper writers}, and wikipedia corresponding to \texttt{wikipedia editor}.

\section{Exeriment Settings}

\subsection{Details of Baselines}
\label{sec:appendix_baseline}
\fakeparagraph{SeqXGPT} based on convolution and self-attention networks, utilizes log probability lists from white-box LLMs as features for sentence-level AIGT detection, has shown a great performance in both sentence and document-level detection challenges but also exhibits strong generalization capabilities.

\fakeparagraph{DeBERTa} improves the BERT\citep{devlin2018bert} and RoBERTa\citep{liu2019roberta} models using disentangled attention and enhanced mask decoder. And it can perform better in classification tasks as well as sequence labeling tasks.

\fakeparagraph{DetectGPT} is a zero-shot whitebox detection method that utilizes log probabilities computed by the model of interest. Based on the hypothesis that minor rewrites of model-generated text tend to have lower log probability under the model than the original sample.

\fakeparagraph{DetectLLM} propose two methods for detection, which are Log-Likelihood Log-Rank ratio (LRR) and Normalized perturbed log-rank (NPR), the methods extensively exploit the potential of the log-rank information.

\fakeparagraph{Fast-DetectGPT} is an optimized zero-shot detector, which substitutes the previous perturbation step with a more efficient sampling step. It also introduces the concept of conditional probability curvature to elucidate discrepancies in word choices between LLMs and humans within a given context.

\fakeparagraph{GLTR} assume that systems overgenerate from a limited subset of the true distribution of natural language, and it utilizes a suite of metric-based methods to aid in human identification including probability, rank, and entropy, 

\subsection{Adaption Details of Baseline Methods}
\label{sec:appendix_adaptions}
Some of the baseline methods are not designed with compatibility with word-level detection under our tasks. Hence, we have adapted some of the baseline methods to sequence labeling.

For DeBERTa \citep{he2023debertav3}, we train it on the sequence labeling task and get the labels of tokens. 

For metric-based methods with perturbation, we apply a new method to perturb the text and calculate the metric. Because the text in \dataset is long, using the T5 \citep{raffel2020exploring} for perturbation would have caused a significant time expense. Instead, we employed the synonym attack method proposed in \citep{dugan-etal-2024-raid} for perturbation and calculated metrics of each word for classification as GLTR \citep{gehrmann-etal-2019-gltr}. 
For LRR in DetectLLM \citep{su2023detectllm}, the same method is applied for metric calculation. The base models used to calculate the metrics in the metrics-based methods are Llama-3-8B, GPT-J-6B, GPT-Neo-2.7B, and GPT-2-XL.

\subsection{Settings For Finetune-Based Methods}
To get the best performance of the method, we test the hyperparameters used in the training process of the finetune-based methods. We use the following hyperparameters in final:

For DeBERTa, we set the warmup ratio of {0.2}, the learning rate of {5e-5}, weight decay of {0}, max position embeddings of {2048}, batch size of {2}. We set the optimizer as AdamW, $\beta_1$ of AdamW is {0.9}, $\beta_2$ of AdamW is {0.999}, $\epsilon$ of AdamW is {1e-8}. We set the scheduler type as {linear}.

For SeqXGPT, we set the warmup ratio of {0.1}, the learning rate of {5e-5}, weight decay of {0.1}, batch size of {8}. We set the optimizer as AdamW, $\beta_1$ of AdamW is {0.9}, $\beta_2$ of AdamW is {0.98}, $\epsilon$ of AdamW is {1e-8}. We set the scheduler type as {linear}.

\label{sec:appendix_adapt}
\section{Additional Results}
\label{sec_appendix_result}
\subsection{Precision and Recall For Classification}
\label{sec_appendix_pr}
We also have the precision and recall for the classification of two text sources, AI and human, respectively. These results can reflect the detailed performance difference of the methods for different text sources, as an addition to the F1 score. The result is shown in \tabref{tab:res_more_1}, \tabref{tab:res_more_2}, \tabref{tab:res_more_3}, \tabref{tab:res_more_4}, and \tabref{tab:res_more_5}. 
\subsection{Absolute Values of Metrics}
\label{sec_appendix_ave}
Besides, since there is a difference in AUC ROC obtained for the metric-based methods between sentence-level detection and word-level detection,
we report the average absolute values of metrics in HWT and MGT to show the mathematical distribution of coauthored texts. The result is shown in \tabref{tab:res_more_6} and \tabref{tab:res_more_7}. 
\subsection{More Backbone Models}
\label{sec_appendix_backbone}
we apply more pre-trained models in the detection as encoders including RoBERTa \citep{liu2019roberta}, XLNet \citep{yang2019xlnet}, ELECTRA \citep{ClarkLLM20}. The results of the different backbone models in the in-domain experiments are presented in the \tabref{tab:res_more_8}, which shows that DeBERTa is the best encoder.

\subsection{Sentence-Level Metric-Based Detectors}
\label{sec_appendix_sentence}
We try to directly compute the metrics of sentences for sentence-level detection with metric-based detectors. The results of in-domain experiments are shown in the \tabref{tab:res_more_7}. It shows that there is no significant difference between its performance and that of the majority vote method we adapted.

\begin{table*}[]
\centering
\begin{tabular}{@{\hspace{2mm}}lcccc@{\hspace{2mm}}}
\toprule[2pt]
\textbf{Detector} & Human-P.             & Human-R.             & AI-P.                & AI-R.                \\ \midrule[1pt]
Random            & 0.696                & 0.659                & 0.302                & 0.339                      \\ \midrule[1pt]
\multicolumn{5}{l}{\textbf{Finetune-based Methods}}                                                          \\ \midrule[1pt]
DeBERTa           & {\ul \textbf{0.980}} & {\ul \textbf{0.978}} & {\ul \textbf{0.950}} & {\ul \textbf{0.954}} \\
SeqXGPT           & 0.810                & 0.903                & 0.608                & 0.416                \\ \midrule[1pt]
\multicolumn{5}{l}{\textbf{Metric-based Methods (w/ Perturb)}}                                                \\ \midrule[1pt]
DetectGPT         & 0.697                & {\ul 0.923}          & 0.305                & 0.078                \\
NPR               & {\ul 0.736}          & 0.386                & 0.325                & {\ul 0.681}          \\
Fast-DetectGPT    & 0.724                & 0.638                & {\ul 0.346}          & 0.441                \\ \midrule[1pt]
\multicolumn{5}{l}{\textbf{Metric-based Methods (w/o Perturb)}}                                               \\ \midrule[1pt]
log prob          & 0.759                & 0.298                & 0.326                & 0.783                \\
rank              & 0.747                & 0.352                & 0.327                & 0.726                \\
log rank          & 0.734                & 0.485                & 0.334                & 0.595                \\
entropy           & {\ul 0.762}          & 0.203                & 0.318                & {\ul 0.854}          \\
LRR               & 0.733                & {\ul 0.518}          & {\ul 0.338}          & 0.566                \\ \bottomrule[2pt]
\end{tabular}
\caption{Precision and recall of human class and AI class in IND fine-grained sentence-level MGT detection. '-P’ means Precision and ‘-R’ means Recall. Bold means the overall best performance, and underline means the best performance in the categories. `Random' refers to the result of random prediction.}
\label{tab:res_more_1}
\end{table*}

\begin{table*}[]
\centering
\begin{tabular}{@{\hspace{2mm}}lcccc@{\hspace{2mm}}}
\toprule[2pt]
\textbf{Detector} & Human-P.             & Human-R.             & AI-P.                & AI-R.                \\ \midrule[1pt]
Random            & 0.877                & 0.545                & 0.122                 & 0.453                     \\ \midrule[1pt]
\multicolumn{5}{l}{\textbf{Finetune-based Methods}}                                                          \\ \midrule[1pt]
DeBERTa           & {\ul \textbf{0.957}} & {\ul \textbf{0.961}} & {\ul \textbf{0.714}} & {\ul \textbf{0.693}} \\
SeqXGPT           & 0.896                & 0.996                & 0.606                & 0.046                \\ \midrule[1pt]
\multicolumn{5}{l}{\textbf{Metric-based Methods (w/ Perturb)}}                                                \\ \midrule[1pt]
DetectGPT         & 0.869                & {\ul 0.409}          & 0.117                & 0.561                \\
NPR               & {\ul 0.871}          & 0.509                & 0.116                & {\ul 0.461}          \\
Fast-DetectGPT    & 0.761                & 0.665                & {\ul 0.251}          & 0.349                \\ \midrule[1pt]
\multicolumn{5}{l}{\textbf{Metric-based Methods (w/o Perturb)}}                                               \\ \midrule[1pt]
log prob          & 0.873                & 0.768                & 0.108                & 0.201                \\
rank              & 0.873                & 0.588                & 0.116                & 0.388                \\
log rank          & 0.873                & 0.584                & 0.116                & 0.391                \\
entropy           & {\ul 0.873}          & 0.779                & 0.105                & {\ul 0.185}          \\
LRR               & 0.873                & {\ul 0.748}          & {\ul 0.109}          & 0.220                \\ \bottomrule[2pt]
\end{tabular}
\caption{Precision and recall of human class and AI class in IND fine-grained word-level MGT detection. '-P’ means Precision and ‘-R’ means Recall. Bold means the overall best performance, and underline means the best performance in the categories. `Random' refers to the result of random prediction.}
\end{table*}
\begin{table*}[]
\centering
\scalebox{0.55}{
\begin{tabular}{@{\hspace{2mm}}lcccccccccccccccc@{\hspace{2mm}}}
\toprule[2pt]
\multicolumn{1}{c}{\multirow{2}{*}{Detector}} & \multicolumn{4}{c}{News}                                                                        & \multicolumn{4}{c}{Paper}                                                                       & \multicolumn{4}{c}{Story}                                                                       & \multicolumn{4}{c}{Wikipedia}                                                             \\ \cmidrule(l){2-17} 
\multicolumn{1}{c}{}                          & Human-P.             & Human-R.             & AI-P.                & \multicolumn{1}{c|}{AI-R.} & Human-P.             & Human-R.             & AI-P.                & \multicolumn{1}{c|}{AI-R.} & Human-P.             & Human-R.             & AI-P.                & \multicolumn{1}{c|}{AI-R.} & Human-P.             & Human-R.             & AI-P.                & AI-R.                \\ \midrule[1pt]
random   & 0.572           & 0.660               & 0.421               & 0.334            & 0.687                   & 0.661               & 0.307               & 0.332            & 0.627                   & 0.623               & 0.364               & 0.368            & 0.757                   & 0.672               & 0.241               & 0.326                     \\ \midrule[1pt]
\multicolumn{17}{l}{\textbf{Fine-tune based Methods}}                                                                                                                                                                                                                                                                                                                                                                                           \\ \midrule[1pt]
DeBERTa                                       & {\ul \textbf{0.973}} & {\ul \textbf{0.966}} & {\ul \textbf{0.954}} & {\ul \textbf{0.963}}       & {\ul \textbf{0.979}} & 0.917                & {\ul \textbf{0.839}} & {\ul \textbf{0.955}}       & {\ul \textbf{0.953}} & {\ul \textbf{0.900}} & {\ul \textbf{0.844}} & {\ul \textbf{0.925}}       & {\ul \textbf{0.982}} & 0.952                & {\ul \textbf{0.864}} & {\ul \textbf{0.945}} \\
SeqXGPT                                       & 0.597                & 0.859                & 0.537                & 0.220                      & 0.818                & {\ul \textbf{0.993}} & 0.242                & 0.009                      & 0.681                & 0.833                & 0.522                & 0.320                      & 0.892                & {\ul \textbf{0.997}} & 0.321                & 0.013                \\ \midrule[1pt]
\multicolumn{17}{l}{\textbf{Metric-based Methods(w)}}                                                                                                                                                                                                                                                                                                                                                                                           \\ \midrule[1pt]
DetectGPT                                     & 0.566                & {\ul 0.900}          & 0.330                & 0.067                      & 0.686                & {\ul 0.932}          & 0.268                & 0.055                      & 0.689                & 0.228                & 0.385                & {\ul 0.824}                & 0.757                & {\ul 0.995}          & 0.191                & 0.004                \\
NPR                                           & {\ul 0.628}          & 0.412                & 0.458                & {\ul 0.670}                & 0.690                & 0.859                & 0.319                & 0.146                      & {\ul 0.698}          & 0.550                & {\ul 0.436}          & 0.594                      & 0.790                & 0.377                & {\ul 0.261}          & 0.688                \\
Fast-DetectGPT                                & 0.612                & 0.621                & {\ul 0.477}          & 0.468                      & {\ul 0.712}          & 0.665                & {\ul 0.353}          & {\ul 0.404}                & 0.672                & {\ul 0.637}          & 0.431                & 0.469                      & {\ul 0.795}          & 0.345                & {\ul 0.261}          & {\ul 0.722}          \\ \midrule[1pt]
\multicolumn{17}{l}{\textbf{Metric-based Methods(w/o)}}                                                                                                                                                                                                                                                                                                                                                                                         \\ \midrule[1pt]
log prob                                      & 0.670                & 0.322                & 0.462                & 0.786                      & {\ul 0.733}          & 0.355                & 0.333                & 0.714                      & 0.734                & 0.476                & 0.441                & 0.705                      & 0.819                & 0.196                & 0.256                & 0.865                \\
rank                                          & 0.647                & 0.345                & 0.457                & 0.745                      & 0.712                & {\ul 0.525}          & 0.335                & 0.529                      & 0.716                & 0.517                & 0.441                & 0.650                      & 0.803                & 0.298                & 0.261                & 0.773                \\
log rank                                      & 0.628                & 0.505                & 0.471                & 0.596                      & 0.719                & 0.464                & 0.335                & 0.598                      & {\ul 0.749}          & 0.393                & 0.428                & {\ul 0.776}                & {\ul 0.820}          & 0.168                & 0.254                & {\ul 0.884}          \\
entropy                                       & {\ul 0.711}          & 0.218                & 0.454                & {\ul 0.880}                & 0.731                & 0.214                & 0.321                & {\ul 0.825}                & 0.731                & 0.372                & 0.417                & 0.767                      & 0.809                & 0.158                & 0.252                & {\ul 0.884}          \\
LRR                                           & 0.628                & {\ul 0.543}          & {\ul 0.478}          & 0.566                      & 0.720                & 0.506                & {\ul 0.340}          & 0.563                      & 0.680                & {\ul 0.728}          & {\ul 0.472}          & 0.415                      & 0.797                & {\ul 0.459}          & {\ul 0.273}          & 0.636                \\ \bottomrule[2pt]
\end{tabular}
}
\caption{Precision and recall of human class and AI class in out-of-domain sentence-level MGT detection. -P’ means Precision and ‘-R’ means Recall. Bold means the overall best performance, and underline means the best performance in the categories. `Random' refers to the result of random prediction.}
\label{tab:res_more_2}
\end{table*}
\begin{table*}[]
\centering
\scalebox{0.55}{
\begin{tabular}{@{\hspace{2mm}}lcccccccccccccccc@{\hspace{2mm}}}
\toprule[2pt]
\multicolumn{1}{c}{\multirow{2}{*}{Detector}} & \multicolumn{4}{c}{News}                                                                        & \multicolumn{4}{c}{Paper}                                                                       & \multicolumn{4}{c}{Story}                                                                       & \multicolumn{4}{c}{Wikipedia}                                                             \\ \cmidrule(l){2-17} 
\multicolumn{1}{c}{}                          & Human-P.             & Human-R.             & AI-P.                & \multicolumn{1}{c|}{AI-R.} & Human-P.             & Human-R.             & AI-P.                & \multicolumn{1}{c|}{AI-R.} & Human-P.             & Human-R.             & AI-P.                & \multicolumn{1}{c|}{AI-R.} & Human-P.             & Human-R.             & AI-P.                & AI-R.                \\ \midrule[1pt]
random & 0.797           & 0.546               & 0.204               & 0.455            & 0.884                   & 0.545               & 0.116               & 0.455            & 0.795                   & 0.546               & 0.206               & 0.456            & 0.909                   & 0.546               & 0.091               & 0.455                     \\ \midrule[1pt]
\multicolumn{17}{l}{\textbf{Fine-tune based Methods}}                                                                                                                                                                                                                                                                                                                                                                                           \\ \midrule[1pt]
DeBERTa                                       & {\ul \textbf{0.927}} & 0.935                & {\ul \textbf{0.735}} & {\ul \textbf{0.710}}       & {\ul \textbf{0.945}} & 0.956                & {\ul \textbf{0.629}} & {\ul \textbf{0.575}}       & {\ul \textbf{0.925}} & 0.940                & {\ul \textbf{0.751}} & {\ul \textbf{0.706}}       & {\ul \textbf{0.965}} & 0.961                & {\ul \textbf{0.624}} & {\ul \textbf{0.649}} \\
SeqXGPT                                       & 0.797                & {\ul \textbf{0.997}} & 0.424                & 0.007                      & 0.934                & {\ul \textbf{0.999}} & 0.002                & 0.000                      & 0.801                & {\ul \textbf{0.998}} & 0.436                & 0.007                      & 0.960                & {\ul \textbf{1.000}} & 0.303                & 0.000                \\ \midrule[1pt]
\multicolumn{17}{l}{\textbf{Metric-based Methods(w)}}                                                                                                                                                                                                                                                                                                                                                                                           \\ \midrule[1pt]
DetectGPT                                     & 0.779                & 0.382                & 0.192                & {\ul 0.576}                & 0.874                & 0.379                & 0.110                & {\ul 0.583}                & 0.797                & 0.479                & 0.208                & {\ul 0.528}                & 0.901                & 0.417                & 0.084                & {\ul 0.540}          \\
NPR                                           & 0.787                & 0.504                & 0.194                & 0.466                      & 0.877                & 0.492                & 0.108                & 0.471                      & {\ul 0.800}          & 0.569                & {\ul 0.212}          & 0.449                      & 0.904                & 0.523                & 0.085                & 0.444                \\
Fast-DetectGPT                                & {\ul 0.799}          & {\ul 0.654}          & {\ul 0.208}          & 0.355                      & {\ul 0.885}          & {\ul 0.666}          & {\ul 0.117}          & 0.337                      & 0.799                & {\ul 0.642}          & {\ul 0.212}          & 0.373                      & {\ul 0.911}          & {\ul 0.668}          & {\ul 0.093}          & 0.342                \\ \midrule[1pt]
\multicolumn{17}{l}{\textbf{Metric-based Methods(w/o)}}                                                                                                                                                                                                                                                                                                                                                                                         \\ \midrule[1pt]
log prob                                      & {\ul 0.790}          & 0.759                & 0.183                & 0.211                      & {\ul 0.881}          & {\ul 0.807}          & 0.101                & 0.165                      & 0.797                & 0.758                & 0.214                & 0.255                      & {\ul 0.906}          & 0.739                & 0.081                & 0.232                \\
rank                                          & {\ul 0.790}          & 0.592                & {\ul 0.194}          & 0.384                      & 0.879                & 0.583                & {\ul 0.108}          & 0.387                      & 0.799                & 0.644                & 0.213                & 0.374                      & {\ul 0.906}          & 0.596                & {\ul 0.086}          & 0.382                \\
log rank                                      & {\ul 0.790}          & 0.590                & {\ul 0.194}          & {\ul 0.386}                & 0.879                & 0.577                & {\ul 0.108}          & {\ul 0.392}                & {\ul 0.800}          & 0.643                & 0.214                & {\ul 0.376}                & {\ul 0.906}          & 0.594                & {\ul 0.086}          & {\ul 0.383}          \\
entropy                                       & {\ul 0.790}          & {\ul 0.794}          & 0.178                & 0.175                      & 0.880                & 0.771                & 0.102                & 0.199                      & 0.796                & {\ul 0.876}          & {\ul 0.216}          & 0.133                      & 0.905                & {\ul 0.773}          & 0.077                & 0.189                \\
LRR                                           & {\ul 0.790}          & 0.745                & 0.184                & 0.225                      & 0.880                & 0.752                & 0.102                & 0.216                      & 0.798                & 0.791                & {\ul 0.216}          & 0.223                      & {\ul 0.906}          & 0.724                & 0.081                & 0.245                \\ \bottomrule[2pt]
\end{tabular}
}
\caption{Precision and recall of human class and AI class in out-of-domain word-Level MGT detection. '-P’ means Precision and ‘-R’ means Recall. Bold means the overall best performance, and underline means the best performance in the categories. `Random' refers to the result of random prediction.}
\label{tab:res_more_3}
\end{table*}
\begin{table*}[]
\centering
\scalebox{0.55}{
\begin{tabular}{@{\hspace{2mm}}lcccccccccccccccc@{\hspace{2mm}}}
\toprule[2pt]
\multicolumn{1}{c}{\multirow{2}{*}{Detector}} & \multicolumn{4}{c}{Llama3}                                                                      & \multicolumn{4}{c}{Mixtral}                                                                     & \multicolumn{4}{c}{GPT-4o}                                                                      & \multicolumn{4}{c}{GPT-4o-mini}                                                           \\ \cmidrule(l){2-17} 
\multicolumn{1}{c}{}                          & Human-P.             & Human-R.             & AI-P.                & \multicolumn{1}{c|}{AI-R.} & Human-P.             & Human-R.             & AI-P.                & \multicolumn{1}{c|}{AI-R.} & Human-P.             & Human-R.             & AI-P.                & \multicolumn{1}{c|}{AI-R.} & Human-P.             & Human-R.             & AI-P.                & AI-R.                \\ \midrule[1pt]
random  & 0.707            & 0.658                & 0.285                  & 0.332          & 0.700                 & 0.658                & 0.296                  & 0.337          & 0.689                 & 0.659                & 0.312                  & 0.343          & 0.688                 & 0.661                & 0.313                  & 0.340                     \\ \midrule[1pt]
\multicolumn{17}{l}{\textbf{Fine-tune based Methods}}                                                                                                                                                                                                                                                                                                                                                                                           \\ \midrule[1pt]
DeBERTa                                       & {\ul \textbf{0.988}} & {\ul \textbf{0.984}} & {\ul \textbf{0.961}} & {\ul \textbf{0.970}}       & {\ul \textbf{0.966}} & {\ul \textbf{0.977}} & {\ul \textbf{0.944}} & {\ul \textbf{0.918}}       & {\ul \textbf{0.950}} & {\ul \textbf{0.968}} & {\ul \textbf{0.926}} & {\ul \textbf{0.886}}       & {\ul \textbf{0.986}} & {\ul \textbf{0.974}} & {\ul \textbf{0.945}} & {\ul \textbf{0.969}} \\
SeqXGPT                                       & 0.797                & 0.912                & 0.624                & 0.387                      & 0.805                & 0.912                & 0.962                & 0.395                      & 0.807                & 0.904                & 0.575                & 0.376                      & 0.803                & 0.904                & 0.598                & 0.391                \\ \midrule[1pt]
\multicolumn{17}{l}{\textbf{Metric-based Methods(w)}}                                                                                                                                                                                                                                                                                                                                                                                           \\ \midrule[1pt]
DetectGPT                                     & 0.708                & {\ul 0.925}          & 0.267                & 0.067                      & 0.702                & {\ul 0.924}          & 0.301                & 0.077                      & 0.692                & {\ul 0.921}          & {\ul 0.351}          & 0.094                      & 0.686                & {\ul 0.923}          & 0.296                & 0.071                \\
NPR                                           & {\ul 0.766}          & 0.380                & 0.321                & {\ul 0.717}                & {\ul 0.751}          & 0.395                & 0.328                & {\ul 0.692}                & 0.695                & 0.854                & 0.344                & 0.169                      & {\ul 0.710}          & 0.426                & 0.328                & {\ul 0.618}          \\
Fast-DetectGPT                                & 0.753                & 0.646                & {\ul 0.358}          & 0.482                      & 0.743                & 0.550                & {\ul 0.344}          & 0.554                      & {\ul 0.705}          & 0.634                & 0.338                & {\ul 0.414}                & 0.704                & 0.627                & {\ul 0.338}          & 0.420                \\ \midrule[1pt]
\multicolumn{17}{l}{\textbf{Metric-based Methods(w/o)}}                                                                                                                                                                                                                                                                                                                                                                                         \\ \midrule[1pt]
log prob                                      & 0.798                & 0.318                & 0.325                & 0.803                      & 0.777                & 0.306                & 0.327                & 0.794                      & 0.725                & 0.290                & 0.325                & 0.757                      & 0.732                & 0.330                & 0.332                & 0.733                \\
rank                                          & 0.784                & 0.346                & 0.324                & 0.767                      & 0.765                & 0.360                & 0.330                & 0.740                      & 0.718                & 0.344                & 0.326                & 0.702                      & 0.722                & 0.357                & 0.330                & 0.697                \\
log rank                                      & {\ul 0.812}          & 0.222                & 0.315                & {\ul 0.875}                & {\ul 0.790}          & 0.232                & 0.322                & {\ul 0.855}                & 0.711                & 0.476                & 0.331                & 0.573                      & 0.711                & 0.491                & 0.334                & 0.561                \\
entropy                                       & 0.783                & 0.187                & 0.305                & 0.874                      & 0.770                & 0.208                & 0.315                & 0.854                      & {\ul 0.743}          & 0.186                & 0.323                & {\ul 0.858}                & {\ul 0.759}          & 0.201                & 0.328                & {\ul 0.860}          \\
LRR                                           & 0.753                & {\ul 0.553}          & {\ul 0.338}          & 0.557                      & 0.742                & {\ul 0.527}          & {\ul 0.339}          & 0.570                      & 0.714                & {\ul 0.512}          & {\ul 0.336}          & 0.547                      & 0.718                & {\ul 0.519}          & {\ul 0.342}          & 0.551                \\ \bottomrule[2pt]
\end{tabular}
}
\caption{Precision and recall of human class and AI class in out-of-model sentence-level MGT detection. '-P’ means
Precision and ‘-R’ means Recall. Bold means the overall best performance, and underline means the best performance in the categories. `Random' refers to the result of random prediction.}
\label{tab:res_more_4}
\end{table*}
\begin{table*}[]
\centering
\scalebox{0.55}{
\begin{tabular}{@{\hspace{2mm}}lcccccccccccccccc@{\hspace{2mm}}}
\toprule[2pt]
\multicolumn{1}{c}{\multirow{2}{*}{Detector}} & \multicolumn{4}{c}{Llama3}                                                                      & \multicolumn{4}{c}{Mixtral}                                                                     & \multicolumn{4}{c}{GPT-4o}                                                                      & \multicolumn{4}{c}{GPT-4o-mini}                                                           \\ \cmidrule(l){2-17} 
\multicolumn{1}{c}{}                          & Human-P.             & Human-R.             & AI-P.                & \multicolumn{1}{c|}{AI-R.} & Human-P.             & Human-R.             & AI-P.                & \multicolumn{1}{c|}{AI-R.} & Human-P.             & Human-R.             & AI-P.                & \multicolumn{1}{c|}{AI-R.} & Human-P.             & Human-R.             & AI-P.                & AI-R.                \\ \midrule[1pt]
random  & 0.841               & 0.546               & 0.159                  & 0.454       & 0.877                 & 0.546                 & 0.124                  & 0.456         & 0.920                 & 0.545                 & 0.081                  & 0.455         & 0.874                 & 0.545                 & 0.127                  & 0.455                     \\ \midrule[1pt]
\multicolumn{17}{l}{\textbf{Fine-tune based Methods}}                                                                                                                                                                                                                                                                                                                                                                                           \\ \midrule[1pt]
DeBERTa                                       & {\ul \textbf{0.951}} & 0.958                & {\ul \textbf{0.769}} & {\ul \textbf{0.740}}       & {\ul \textbf{0.946}} & 0.967                & {\ul \textbf{0.723}} & {\ul \textbf{0.607}}       & {\ul \textbf{0.960}} & 0.969                & {\ul \textbf{0.603}} & {\ul \textbf{0.544}}       & {\ul \textbf{0.950}} & 0.960                & {\ul \textbf{0.703}} & {\ul \textbf{0.649}} \\
SeqXGPT                                       & 0.831                & {\ul \textbf{0.999}} & 0.760                & 0.008                      & 0.879                & {\ul \textbf{0.996}} & 0.485                & 0.027                      & 0.932                & {\ul \textbf{0.995}} & 0.201                & 0.018                      & 0.888                & {\ul \textbf{0.997}} & 0.381                & 0.015                \\ \midrule[1pt]
\multicolumn{17}{l}{\textbf{Metric-based Methods(w)}}                                                                                                                                                                                                                                                                                                                                                                                           \\ \midrule[1pt]
DetectGPT                                     & 0.835                & 0.409                & 0.155                & 0.573                      & 0.871                & 0.421                & 0.119                & {\ul 0.556}                & 0.861                & 0.408                & {\ul 0.118}          & {\ul 0.546}                & {\ul 0.912}          & 0.406                & 0.075                & {\ul 0.552}          \\
NPR                                           & 0.839                & {\ul 0.518}          & 0.158                & 0.476                      & 0.874                & 0.533                & 0.120                & 0.452                      & 0.911                & 0.507                & 0.072                & 0.438                      & 0.861                & 0.501                & {\ul 0.113}          & 0.439                \\
Fast-DetectGPT                                & {\ul 0.847}          & 0.381                & {\ul 0.163}          & {\ul 0.638}                & {\ul 0.895}          & {\ul 0.910}          & {\ul 0.183}          & 0.125                      & {\ul 0.961}          & {\ul 0.929}          & 0.058                & 0.103                      & 0.906                & {\ul 0.889}          & 0.112                & 0.132                \\ \midrule[1pt]
\multicolumn{17}{l}{\textbf{Metric-based Methods(w/o)}}                                                                                                                                                                                                                                                                                                                                                                                         \\ \midrule[1pt]
log prob                                      & 0.838                & 0.668                & 0.154                & 0.319                      & 0.874                & 0.717                & 0.116                & 0.264                      & {\ul 0.916}          & {\ul 0.817}          & 0.065                & 0.145                      & {\ul 0.868}          & {\ul 0.850}          & 0.096                & 0.111                \\
rank                                          & {\ul 0.841}          & 0.587                & {\ul 0.159}          & 0.413                      & {\ul 0.875}          & 0.607                & {\ul 0.121}          & 0.386                      & 0.913                & 0.587                & {\ul 0.072}          & 0.366                      & 0.863                & 0.588                & {\ul 0.111}          & 0.357                \\
log rank                                      & {\ul 0.841}          & 0.583                & {\ul 0.159}          & {\ul 0.416}                & {\ul 0.875}          & 0.586                & 0.120                & {\ul 0.403}                & 0.913                & 0.582                & {\ul 0.072}          & {\ul 0.369}                & {\ul 0.868}          & 0.584                & {\ul 0.111}          & {\ul 0.360}          \\
entropy                                       & 0.835                & {\ul 0.792}          & 0.136                & 0.173                      & 0.872                & {\ul 0.794}          & 0.106                & 0.174                      & {\ul 0.916}          & 0.778                & 0.069                & 0.188                      & {\ul 0.868}          & 0.782                & 0.106                & 0.178                \\
LRR                                           & 0.837                & 0.713                & 0.150                & 0.269                      & 0.874                & 0.731                & 0.114                & 0.247                      & 0.915                & 0.763                & 0.067                & 0.194                      & 0.867                & 0.766                & 0.103                & 0.186                \\ \bottomrule[2pt]
\end{tabular}
}
\caption{Precision and recall of human class and AI class in out-of-model word-level MGT detection. '-P’ means
Precision and ‘-R’ means Recall. Bold means the overall best performance, and underline means the best performance in the categories. `Random' refers to the result of random prediction.}
\label{tab:res_more_5}
\end{table*}
\begin{table*}[]
\centering
\scalebox{0.9}{
\begin{tabular}{@{\hspace{2mm}}ccccccccc@{\hspace{2mm}}}
\toprule
\multicolumn{1}{c|}{Set}    & \multicolumn{4}{c|}{Train}                                 & \multicolumn{4}{c}{Test}              \\
\multicolumn{1}{c|}{Metric} & log prob & rank  & log rank & \multicolumn{1}{c|}{entropy} & log prob & rank  & log rank & entropy \\ \midrule
Average Value in AI         & -3.3457  & 1,288 & 1.7461   & 2.6425                       & -3.3367  & 1,269 & 1.7403   & 2.6389  \\
Average Value in Human      & -3.4613  & 1,344 & 1.8232   & 2.6967                       & -3.4557  & 1,342 & 1.8198   & 2.6936  \\ \bottomrule
\end{tabular}
}
\caption{Average value of metrics in the training set and test set in word-level.}
\label{tab:res_more_6}
\end{table*}
\begin{table*}[]
\centering
\scalebox{0.9}{
\begin{tabular}{@{\hspace{2mm}}ccccccccc@{\hspace{2mm}}}
\toprule
\multicolumn{1}{c|}{Set}    & \multicolumn{4}{c|}{Train}                                 & \multicolumn{4}{c}{Test}              \\
\multicolumn{1}{c|}{Metric} & log prob & rank  & log rank & \multicolumn{1}{c|}{entropy} & log prob & rank  & log rank & entropy \\ \midrule
Average Value in AI      & -3.5896  & 1,402 & 1.8894   & 2.7263                       & -3.5774  & 1,381 & 1.8823   & 2.7251  \\
Average Value in Human         & -3.4032  & 1,317 & 1.7871   & 2.6737                       & -3.3981  & 1,312 & 1.7839   & 2.6705  \\ \bottomrule
\end{tabular}
}
\caption{Average value of metrics in the training set and test set in word-level.}
\label{tab:res_more_7}
\end{table*}
\begin{table*}[]
\centering
\begin{tabular}{@{\hspace{2mm}}ccc@{\hspace{2mm}}}
\toprule
Base Model & F1-W. & F1-S. \\ \midrule
DeBERTa    & 0.831 & 0.966 \\
XLNet      & 0.823 & 0.952 \\
ELECTRA    & 0.750 & 0.874 \\
RoBERTa    & 0.412 & 0.942 \\ \bottomrule
\end{tabular}
\caption{Performance of the finetune-based method with different backbone models as encoder under in-domain setting.}
\label{tab:res_more_8}
\end{table*}

\begin{table*}[h]
\centering
\begin{tabular}{@{\hspace{2mm}}lcccc@{\hspace{2mm}}}
\toprule
\textbf{Detector} & F1 Score (vote)       & AUC (vote)            & F1 Score (metric)     & AUC (metric)          \\ \midrule
\multicolumn{5}{l}{\textbf{Metric-based Methods (w/ Perturb)}}                                                \\ \midrule
DetectGPT         & 0.459                & 0.501                & 0.467                & 0.504                \\
NPR               & 0.473                & 0.509                & 0.473                & {\ul 0.539}          \\
Fast-DetectGPT    & {\ul \textbf{0.533}} & {\ul 0.510}          & {\ul \textbf{0.506}} & 0.531                \\ \midrule
\multicolumn{5}{l}{\textbf{Metric-based Methods (w/o Perturb)}}                                               \\ \midrule
log prob          & 0.444                & {\ul \textbf{0.511}} & {\ul 0.477}          & {\ul \textbf{0.546}} \\
rank              & 0.465                & 0.510                & 0.443                & 0.513                \\
log rank          & 0.506                & {\ul \textbf{0.511}} & {\ul 0.477}          & 0.545                \\
entropy           & 0.392                & {\ul \textbf{0.511}} & 0.442                & 0.532                \\
LRR               & {\ul 0.516}          & 0.510                & 0.474                & 0.539                \\ \bottomrule
\end{tabular}
\caption{Performance of sentence-level metric-based detection methods. Metrics are calculated at the sentence level. Bold means the overall best performance, and underline means the best performance in the categories.}
\label{tab:res_more_9}
\end{table*}
\clearpage
\begin{figure*}[t]
  \includegraphics[width=\textwidth]{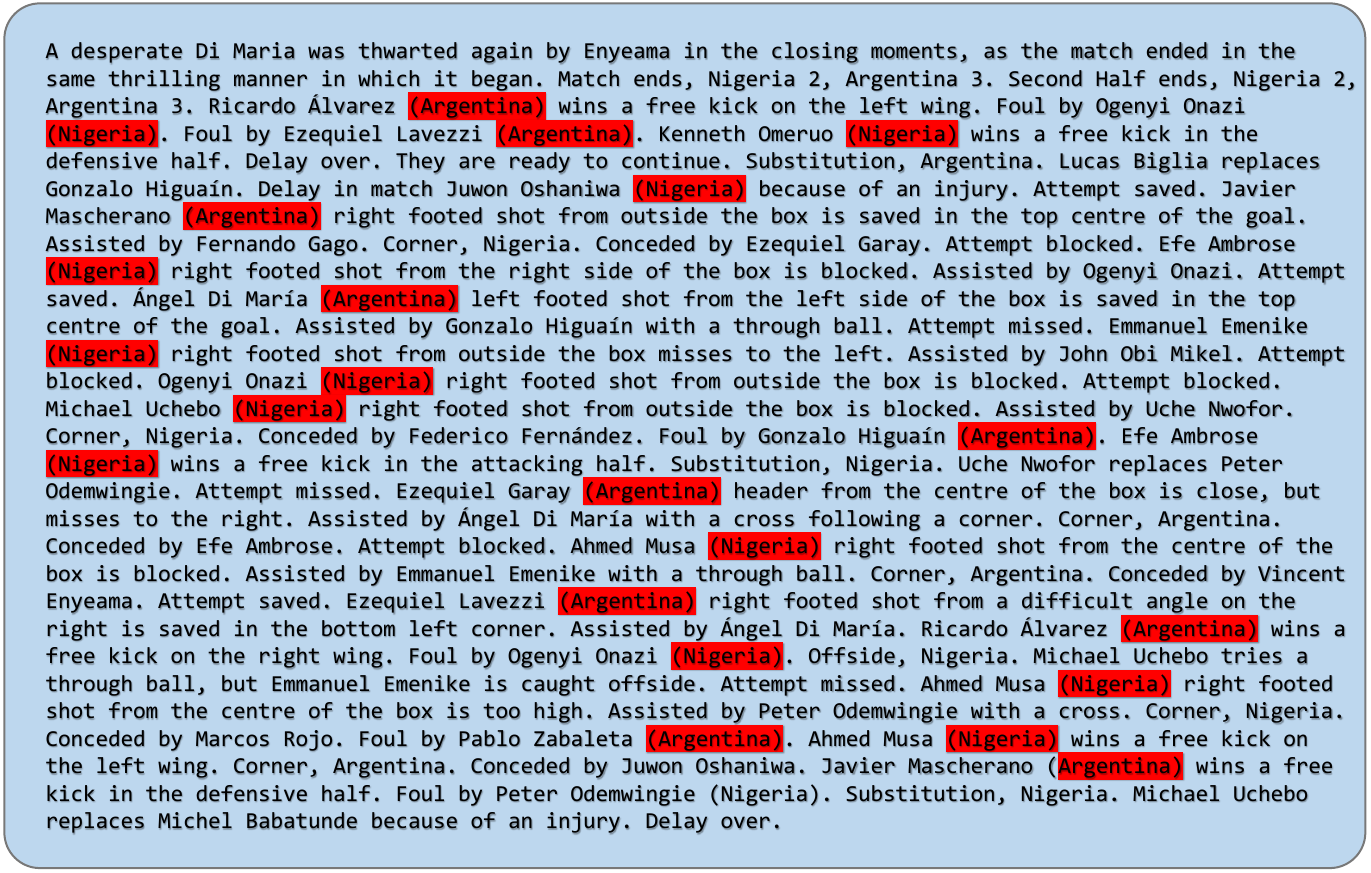}
  \caption{Case of unqualified raw text (Type 1)}
  \label{fig:flt_case_1}
\end{figure*}
\begin{figure*}[t]
  \includegraphics[width=\textwidth]{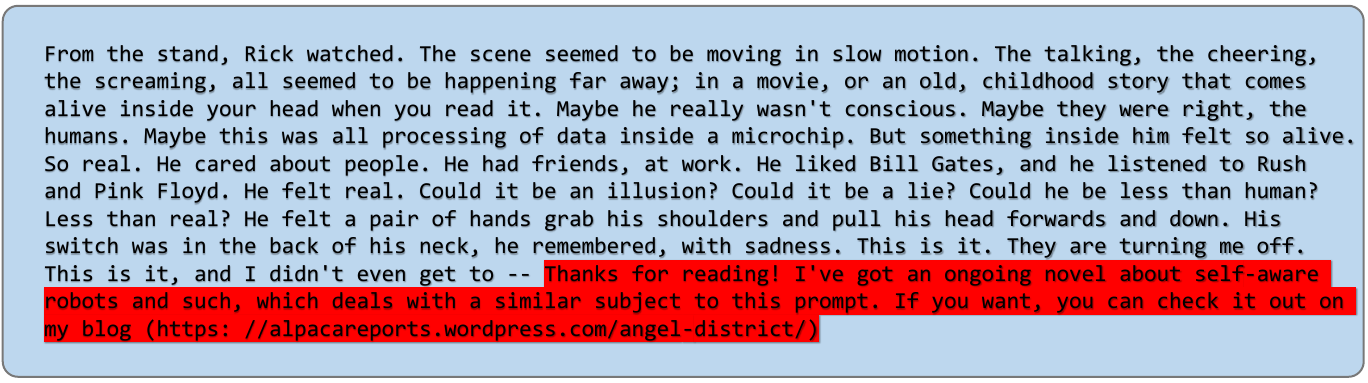}
  \caption{Case of unqualified raw text (Type 2)}
  \label{fig:flt_case_2}
\end{figure*}
\begin{figure*}[t]
  \includegraphics[width=\textwidth]{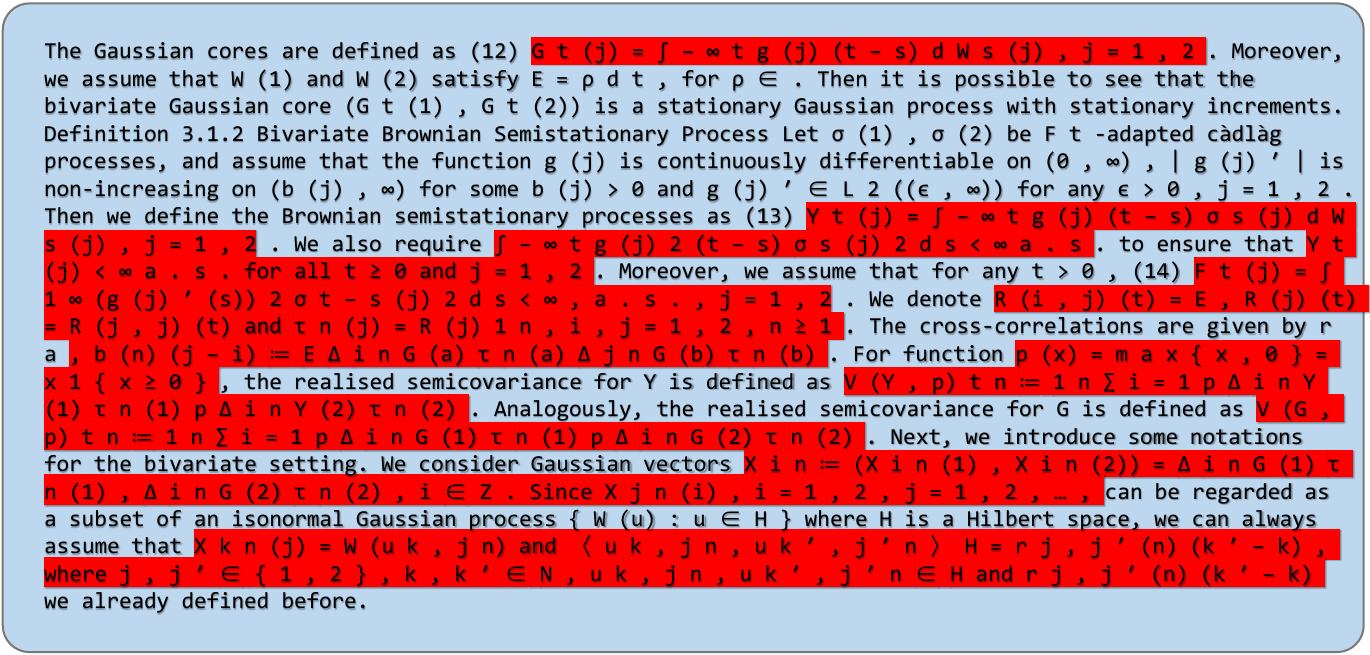}
  \caption{Case of unqualified raw text (Type 3)}
  \label{fig:flt_case_3}
\end{figure*}
\begin{figure*}[t]
  \includegraphics[width=\textwidth]{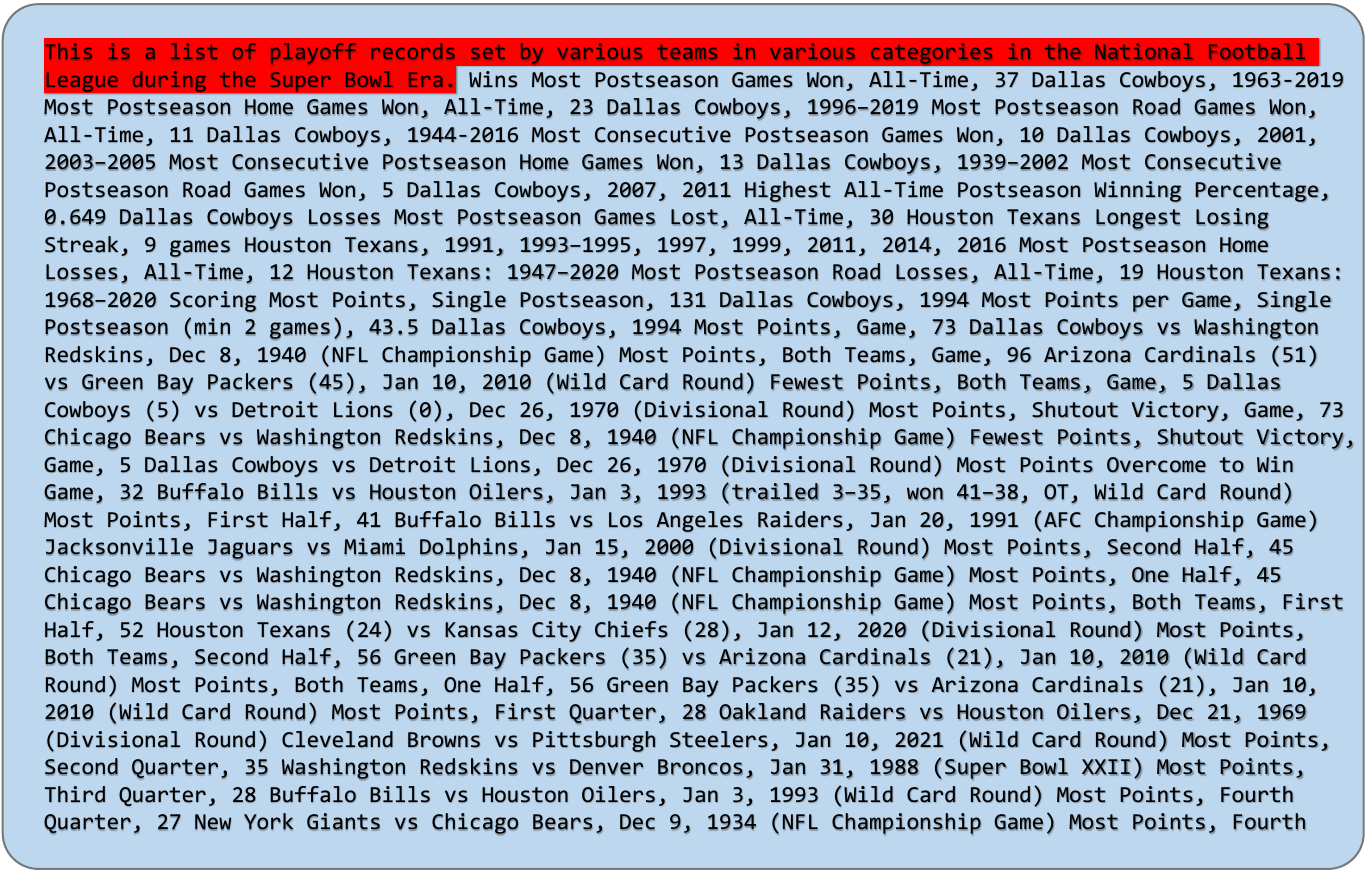}
  \caption{Case of unqualified raw text (Type 4)}
  \label{fig:flt_case_4}
\end{figure*}
\begin{figure*}[t]
  \includegraphics[width=\textwidth]{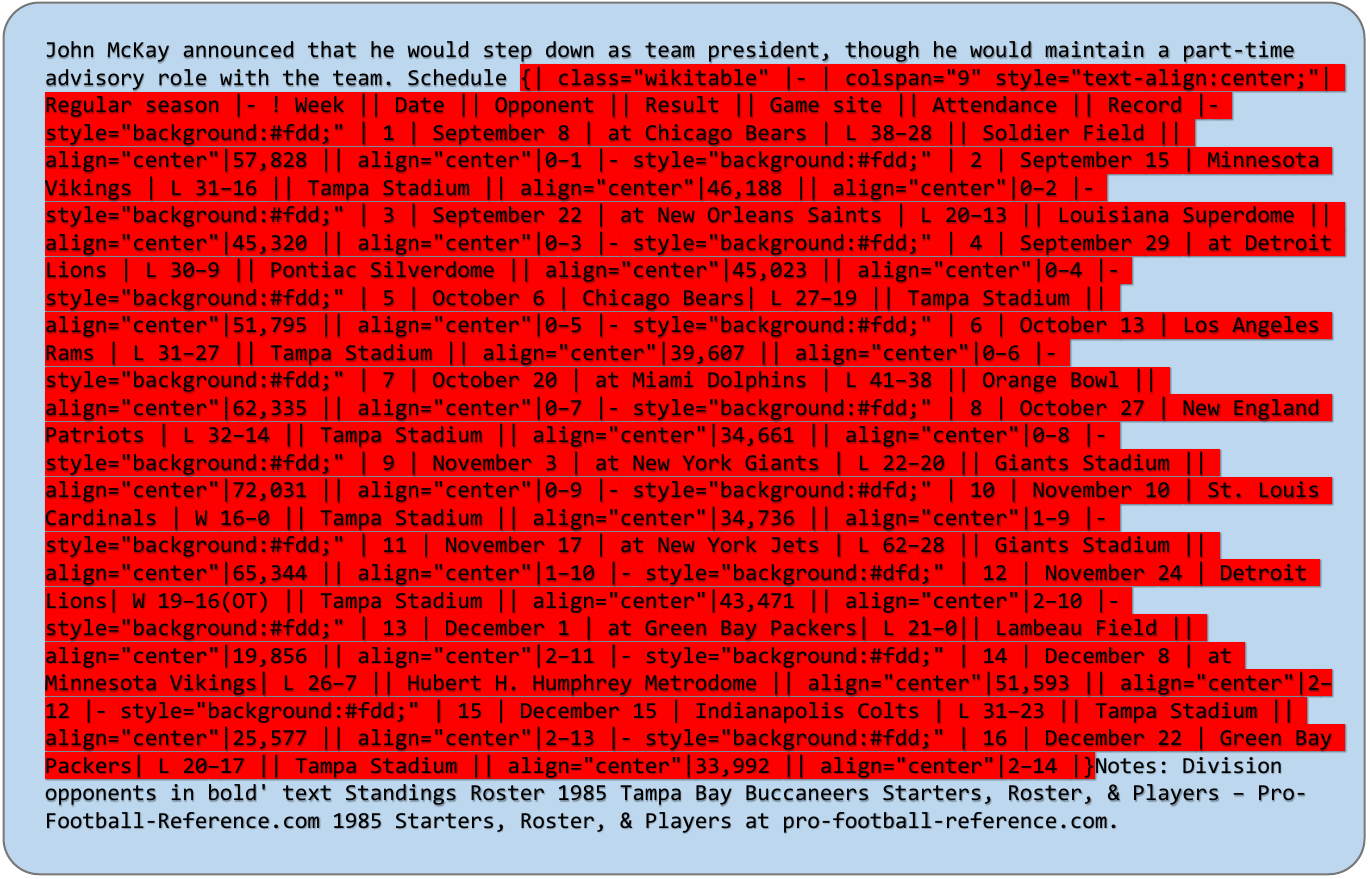}
  \caption{Case of unqualified raw text (Type 5)}
  \label{fig:flt_case_5}
\end{figure*}
\begin{figure*}[t]
  \includegraphics[width=\textwidth]{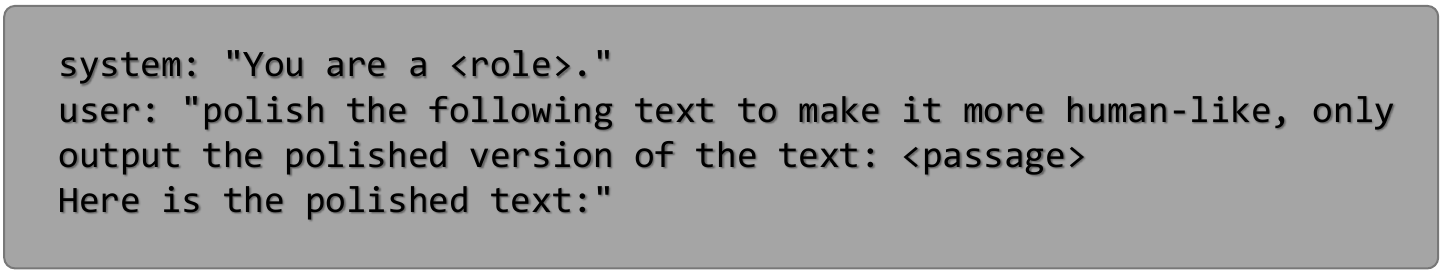}
  \caption{Instruction template for the generation of GPT-4o and GPT-4o-mini.}
  \label{fig:ins_chat_tem}
\end{figure*}
\begin{figure*}[t]
  \includegraphics[width=\textwidth]{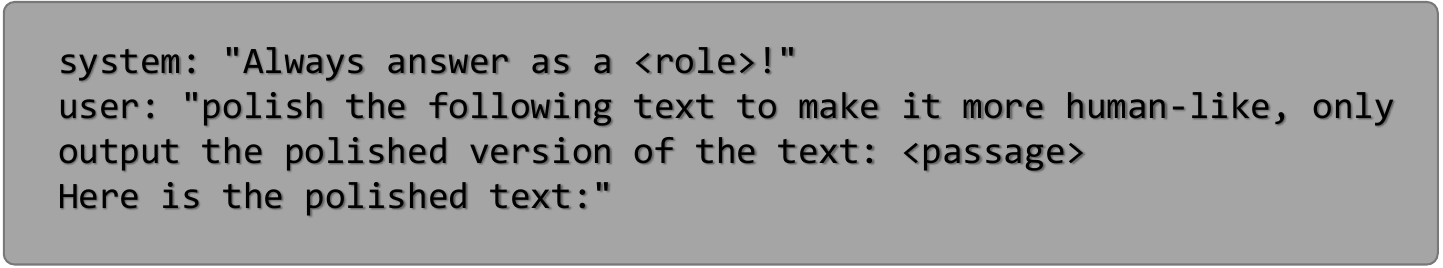}
  \caption{Instruction template for the generation with Llama3.}
  \label{fig:ins_llama_tem}
\end{figure*}
\begin{figure*}[t]
  \includegraphics[width=\textwidth]{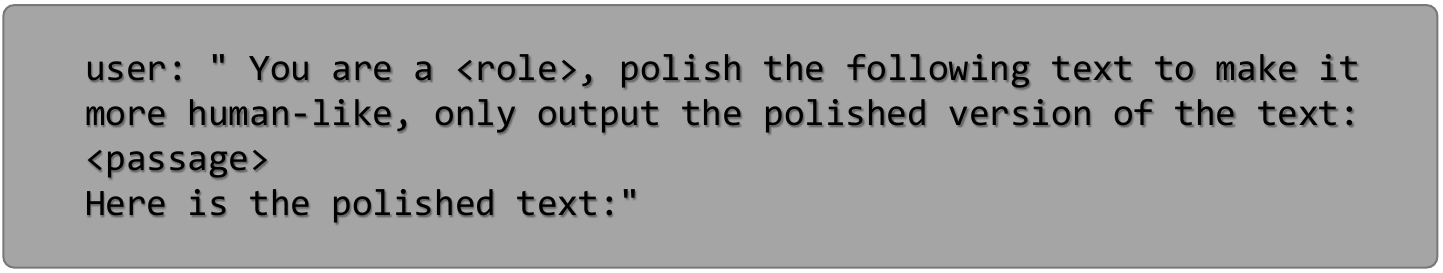}
  \caption{Instruction template for the generation with Mixtral.}
  \label{fig:ins_mixtral_tem}
\end{figure*}
\end{document}